\documentclass[times, review, 10pt]{elsarticle}

\usepackage{amssymb}
\usepackage{amsmath}
\usepackage{microtype}
\usepackage{graphicx}
\usepackage{algorithm}
\usepackage{algorithmic}
\usepackage[switch]{lineno}
\usepackage{graphicx}
\usepackage{array}
\usepackage{caption}
\usepackage{colortbl}
\usepackage{multirow}
\usepackage{booktabs} 
\usepackage{amsmath}
\usepackage{diagbox} 
\usepackage{adjustbox}
\usepackage{pifont}
\usepackage{floatflt}
\usepackage{subcaption}
\usepackage{caption}
\usepackage{colortbl}
\usepackage{multirow}
\usepackage{amsmath}
\usepackage{dblfloatfix} 
\usepackage[utf8]{inputenc} 
\usepackage[T1]{fontenc}    
\usepackage{hyperref}       
\usepackage{url}            
\usepackage{booktabs}       
\usepackage{amsfonts}       
\usepackage{nicefrac}       
\usepackage{microtype}      
\usepackage{xcolor}         
\usepackage{caption}
\usepackage{colortbl}
\usepackage{multirow}
\usepackage{amsmath}

\usepackage{graphicx}
\usepackage{array}
\usepackage{caption}
\usepackage{colortbl}
\usepackage{multirow}
\usepackage{booktabs} 
\usepackage{amsmath}
\usepackage{diagbox} 
\usepackage{adjustbox}
\usepackage{pifont}
\usepackage{wrapfig}
\usepackage{floatflt}
\usepackage{subcaption}
\usepackage{todonotes}
\usepackage{float}

\journal{Nuclear Physics B}

\begin{document}

\begin{frontmatter}

\title{Focal-RegionFace: Generating Fine-Grained Multi-attribute Descriptions for Arbitrarily Selected Face Focal Regions}

\author[label1]{Kaiwen Zheng}
\ead{k.zheng.1@research.gla.ac.uk}
\author[label1]{Junchen Fu}
\ead{j.fu.3@research.gla.ac.uk}
\author[label1]{Songpei Xu}
\ead{s.xu.1@research.gla.ac.uk}
\author[label1]{Yaoqing He}
\ead{heyaoqin1009@gmail.com}
\author[label1]{Joemon M. Jose}
\ead{Joemon.Jose@glasgow.ac.uk}
\author[label3]{Hu Han}
\ead{hanhu@ict.ac.cn}
\author[label2]{Xuri Ge\corref{cor1}}
\cortext[cor1]{Corresponding author.}
\ead{xuri.ge@sdu.edu.cn}

\affiliation[label1]{
    organization={University of Glasgow},
    city={Glasgow},
    country={United Kingdom}
}

\affiliation[label2]{
    organization={Shandong University},
    city={Shandong},
    country={China}
}

\affiliation[label3]{
    organization={Institute of Computing Technology, Chinese Academy of Sciences},
    city={Beijing},
    country={China}
}

\begin{abstract}
In this paper, we introduce an underexplored problem in facial analysis: generating and recognizing multi-attribute natural language descriptions,  containing facial action units (AUs), emotional states, and age estimation, for arbitrarily selected face regions (termed \textbf{\textit{FaceFocalDesc}}). We argue that the system's ability to focus on individual facial areas leads to better understanding and control.
To achieve this capability, we construct a new multi-attribute description dataset for arbitrarily selected face regions, providing rich region-level annotations and natural language descriptions.  Further, we propose a fine-tuned vision-language model based on Qwen2.5-VL, called \textbf{Focal-RegionFace} for facial state analysis, which incrementally refines its focus on localized facial features through multiple progressively fine-tuning stages, resulting in interpretable age estimation, FAU and emotion detection. Experimental results show that Focal-RegionFace achieves the best performance on the new benchmark in terms of traditional and widely used metrics, as well as new proposed metrics. This fully verifies its effectiveness and versatility in fine-grained multi-attribute face region-focal analysis scenarios.
\end{abstract}


\end{frontmatter}
\clearpage

\section{Introduction}

\begin{figure*}[t]
  \centering
  \includegraphics[width=\linewidth]{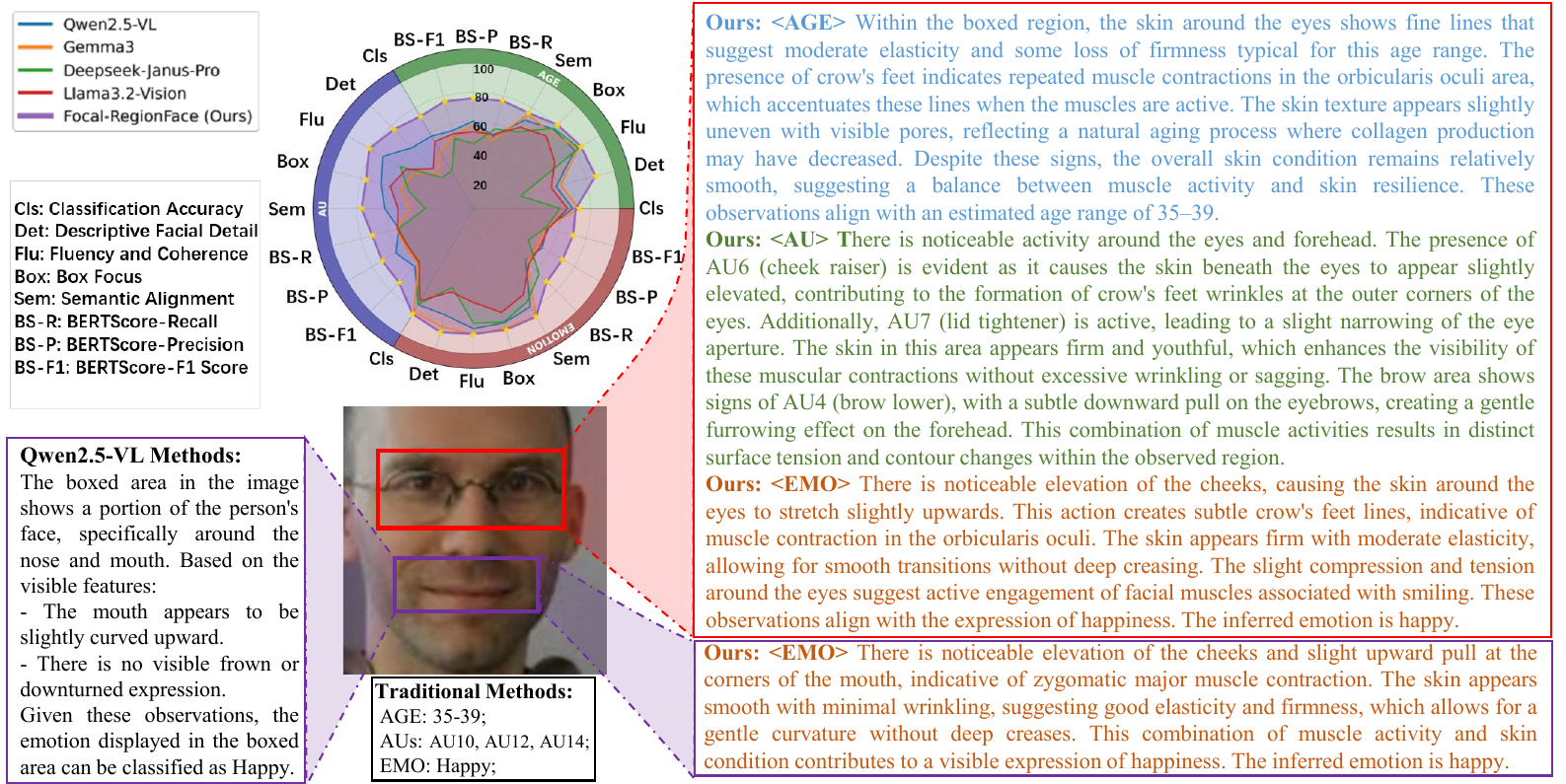}
  \vspace{-1.7em}
  \caption{Comparison of facial state analysis capabilities among mainstream MLLMs and our model achieve superior performance in all NLP metrics. In particular, we show the detailed results of the traditional facial state recognition method, MLLM Qwen2.5-VL and our Focal-RegionFace model. Our Focal-RegionFace model can generate more detailed multi-attribute facial descriptions of arbitrarily selected face regions.}
  \label{fig:motivation}
  \vspace{-1.1em}
\end{figure*}

Human facial analysis is fundamental to vision-language research, underpinning applications in affective computing, medical diagnostics, and human–computer interaction. While traditional methods \cite{MAO2025110951, AGEMethod2018} can predict structured outputs (e.g., AU or emotion categories), these are often limited in interpretability and flexibility. In contrast, natural language descriptions provide more human-aligned and explainable feedback, especially valuable in domains like healthcare and surveillance \cite{ZHAO2026112009, healthcare2025}. 
 Most existing works \cite{wang2025fs, WANG2024110311} focus on global-level face descriptions, while others \cite{wang2025,narayan2025, zheng2025multimodalrepresentationlearningtechniques} explore fine-grained attribute question answering, neglecting the need for localized, fine-grained focal understanding. In practice, users frequently care more about localized facial states, e.g., wrinkle conditions around the eyes or mouth, for tasks like cosmetic or medical recommendation, highlighting the need for fine-grained, region-aware facial focal analysis. In this study, we present a novel solution for an   underexplored task of facial analysis, i.e. \textit{arbitrarily selected facial region state description generation (\textbf{FaceFocalDesc}).}

 \textbf{The capabilities of \textit{FaceFocalDesc}.} As illustrated in Figure \ref{fig:motivation}, our proposed \textit{FaceFocalDesc} introduces a paradigm shift from mainstream facial analysis methods by enabling multi-attribute fine-grained language descriptions for arbitrary facial regions. On one hand, traditional vision-based models \cite{wu2023lanet,zhang2024general} focus on structured prediction of facial states. For instance, \cite{Ganel2022} directly predicts the age of a given facial image in a black-box manner without any explainable information, lacking credibility \cite{Rudin2019}. 
On the other hand, vision-language models are introduced into facial analysis tasks, aiming to improve interpretability by generating human-readable descriptions of facial states. For instance, VL-FAU \cite{Ge_2024_VL} generates crude rule-based linguistic descriptions for facial action unit (AU) states by integrating linguistic generation branches. Recent advances in multimodal large language models have also led to the development of face-domain models, such as Face-LLaVA \cite{chaubey2025facellava}, Emotion-Llama \cite{cheng2024emotion}, which leverage vision-language pretraining to match facial features with global-level descriptive semantics. 
However, these methods remain fundamentally limited in two aspects. First, they rely solely on global face representations, lacking the ability to process arbitrarily user-defined local regions. Second, they typically address only single-attribute outputs (e.g., emotion classification or captioning) and are unable to perform multi-attribute, region-aware facial state modeling. 

\textbf{The challenges of \textit{FaceFocalDesc}.} Despite the above conceptual advantages, building a controllable and interpretable \textit{FaceFocalDesc} system introduces several non-trivial technical challenges. 
First, unlike global face captioning, where the model can rely on holistic cues, \textit{FaceFocalDesc} should be operated under local information constraints, which often lack the full semantic context. The model must therefore learn to reason based on partial visual signals while still maintaining semantic completeness and linguistic fluency. This demands high-level spatial focal awareness. 
Second, integrating multiple facial understanding tasks, such as action unit detection \cite{2019Li, ROMERO2018}, emotion recognition \cite{cheng2024emotion}, and age estimation \cite{shou2025masked}, into a unified language generation framework is non-trivial. These tasks have inherently different semantic structures and visual correlates, and naively combining them can lead to either fragmented or overly generic descriptions. 
Third, existing large-scale datasets for facial description are generally global, sparse, and task-specific, lacking annotations for region-specific, multi-task language outputs. This scarcity of data presents a bottleneck for training and evaluating \textit{FaceFocalDesc}.

\textbf{The proposed method -- \textit{Focal-RegionFace}.} To address the above challenges, we propose a new Focal-RegionFace framework based on a widely-used Qwen2.5-VL model \cite{Qwen2.5-VL} for the new facial analysis paradigm \textit{FaceFocalDesc}, enabling fine-grained, multi-attribute language descriptions for arbitrarily selected facial regions. Focal-RegionFace aims to move beyond global face captioning towards region-aware, controllable, and semantically rich understanding. 

Specifically, we first construct a new benchmark dataset tailored to \textit{FaceFocalDesc}, which includes region-level fine-grained multi-attribute annotations and corresponding multi-attribute labels. This dataset provides the necessary supervised fine-tuning for the pre-trained foundation MLLM \cite{fu2025efficient} to learn spatially grounded, multi-attribute language information. 

After that, we propose a four-stage progressive fine-tuning strategy for Focal-RegionFace. We begin by fine-tuning the base Qwen2.5-VL model on global facial attribute recognition tasks, equipping it with basic facial perception capabilities. Next, we introduce region-guided captioning using full-face images with randomly annotated bounding boxes, allowing the model to learn initial spatial focus and region-aware language generation. To further enhance regional focal precision, we employ masked region fine-tuning, where only the selected facial region remains visible, forcing the model to align language solely with localized visual content. Finally, we leverage the rich region-level descriptions to further fine-tune the model for explicit multi-attribute classification, enhancing its ability to predict AUs, emotions, and age. This progressive design effectively builds strong spatial reasoning and multi-attribute alignment into the model, enabling fine-grained and interpretable facial analysis at arbitrary locations.

\textbf{The main contributions of this paper are as follows:} 
\begin{itemize}
    \item We present a new and important face analysis task, i.e. face region-focal multi-attribute description generation from arbitrarily selected regions (named \textbf{\textit{FaceFocalDesc}}). 
    
    \item We propose a novel multi-stage fine-tuning method based on the Qwen2.5-VL framework for generating region-focused face descriptions, called \textbf{Focal-RegionFace}. A face region can be arbitrarily selected and Focal-RegionFace can create the generation of attribute descriptions including action units, emotions, and age, as well as their corresponding category recognition. 
    \item We construct a new benchmark for \textit{FaceFocalDesc}'s training and evaluation, containing multi-attribute region-level facial state descriptions and corresponding attribute labels. 
    \item In addition to traditional recognition and NLP evaluation metrics, we further propose a new and practical evaluation method for \textit{FaceFocalDesc} based on pre-trained MLLMs, including classification accuracy, detail description ability, fluency and naturalness, local focus, and semantic relevance of the generated descriptions.
    
\end{itemize}
Extensive experiments on the new \textit{FaceFocalDesc} benchmark validate the motivation and effectiveness of our proposed \textbf{Focal-RegionFace} model, facilitating future research of fine-grained interactive face state analysis. Compared with the mainstream MLLMs, such as Qwen2.5-VL, Deepseek-Janus-Pro \cite{chen2025janus} and Llama3.2-Vision \cite{lee2025efficient}, our proposed model achieves the best performance in both generation and recognition, tested on open-source and closed-source evaluation models.

\section{Multimodal Face Region-Focal Dataset}
As shown in Figure \ref{fig:motivation}, although traditional face datasets (e.g., BP4D \cite{ZHANG_2014_BP4D}, AffectNet \cite{Mollahosseini_2019_AffectNet}, UTKFace \cite{zhifei2017cvpr}, etc.) have driven progress in face analysis tasks, there are three main limitations: (1) a focus on black-box tasks (e.g., AU and emotion recognition) with limited interpretability, such as reasoning based on skin texture; (2) interpretability-focused datasets like MERR \cite{chaubey2025facellava} and FaceInstruct-1M \cite{cheng2024emotion} provide global descriptions but lack annotations for arbitrary facial areas; (3) 
 few datasets offer multi-attribute annotations (AU, emotion, age) for fine-grained facial ROIs simultaneously \cite{2001ROI}.

To address these gaps, we introduce the Multimodal Face Region-Focal dataset (MFRF) for the \textit{FaceFocalDesc} task. It supports fine-grained, ROI-centered analysis across AU, emotion, and age, with rich linguistic descriptions to enable interactive and region-aware facial understanding.

\noindent \textbf{Data Collection.} To enable high-quality region-focal face description annotation, we construct a new benchmark integrating four established multi-attribute datasets: BP4D for AU recognition, Aff-Wild2 \cite{kollias_2023_affwild2} and RAF-DB \cite{Yan_2020_RAFAU} for emotion recognition, and UTKFace for age estimation. For the age task, original age labels are remapped into 12 ranges ([0–4], [5–9], …, [50–59], 60+) to reflect gradual facial changes \cite{Ahn2024, Trojahn2015}, while AU and emotion labels remain unchanged.

After filtering redundancy and low-quality samples, we obtain 10,000 images (3,000 from BP4D, 2,000 from Aff-Wild2, and 5,000 from UTKFace), each annotated with attribute labels. For each image, 12 face regions of varied sizes are selected based on facial landmarks \cite{landmark2013} to ensure at least 80\% overlap with key facial areas. Each region is annotated by GPT-4o \cite{ray_2023_chatgpt} using attribute-driven prompts, followed by manual refinement. This process yields 120,000 region-focal face images with fine-grained multi-attribute annotations. Additionally, 60,000 image–description pairs are constructed for multi-attribute fine-tuning.

For comprehensive evaluation, the test set includes 1,000 images (300 from BP4D, 200 from RAF-DB, and 500 from UTKFace), each with 12 random regions, resulting in 12,000 region-level samples in total. The landmark-based region fusion strategy further supports multi-region joint description and serves as prior knowledge for multi-attribute recognition fine-tuning (see Method, Stage IV).

\noindent \textbf{Annotation Strategy.} Unlike conventional global-level facial analysis, our approach introduces region-focal descriptions that explicitly connect structured annotations with interpretable model reasoning. We annotate facial AUs, emotions, and age within randomly selected ROIs, emphasizing localized muscle movements, age-related skin cues, and expressions restricted to the boxed area.

The MFRF prompt design follows three principles: Contextual Focus, Region Constraint, and Structured Generation (details in Appendices). Contextual Focus instructs GPT-4o to act as an attribute expert, attending to fine-grained textures and muscular activity within the ROI. Region Constraint enforces exclusion of out-of-box information and alignment with ground-truth labels for spatial–semantic accuracy. Structured Generation ensures coherent paragraph-style outputs that integrate localized visual details with interpretability.

This design yields a high-quality, region-aware benchmark supporting fine-tuning and evaluation of interpretable models for AU, emotion, and age estimation.



\begin{figure*}[t]
  \centering
  \includegraphics[width=\linewidth]{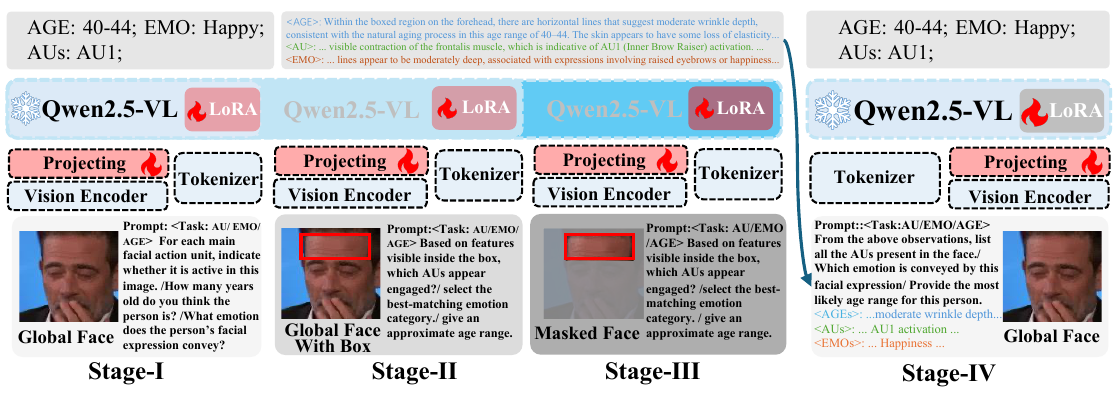}
  \caption{Overview of Focal-RegionFace with multi-stage fine-tuning.  We first perform global face multi-attribute information-aware fine-tuning of Qwen2.5-VL in Stage-I, including age, emotion and AU recognition. Then, we make the model focus on region-focal reasoning in Stage-II and Stage-III in a progressive fine-tuning manner, thus obtaining a Focal-RegionFace MLLM with fine-grained multi-attribute language interpretation. Next, further multimodal inference fine-tuning (Stage-IV) is carried out based on the multi-region visual understanding results, so that the model develops a fine-grained multimodal multi-attribute recognition capability.}
  \label{fig:framework}
\end{figure*}
\section{The Proposed Method} \label{method}
\subsection{Preliminary}
\noindent \textbf{Task Definition.}
\textit{FaceFocalDesc} is formulated as a conditional multi-attribute description generation and recognition task, including action units, emotion, and age, enabling region-aware interpretability. Given a facial image $I$ and an arbitrarily selected region (Region Of Interest, ROI), it could generate fine-grained, multi-attribute natural language descriptions $D_{<AU/EMO/AGE>}$. After that, it can further give the final attribute decisions $P_{<AU/EMO/AGE>}$ with the historical region descriptions ${D_{<AU/EMO/AGE>}}$ as a prompt. This formulation supports both single-turn and history-aware generation modes, facilitating progressive, interpretable facial analysis.

\noindent \textbf{Focal-RegionFace.} To address the above task, we propose Focal-RegionFace in Figure \ref{fig:framework}, a four-stage progressive fine-tuning framework designed to enhance facial region-focal understanding and multi-attribute language generation. Specifically, the framework includes: (Stage I) Global-aware Face Perception, which enables the pre-trained foundation model to acquire comprehensive facial visual representation perception; (Stage II) Region-aware Visual-Language Alignment, which establishes initial capabilities for ROI localization and semantic reasoning; (Stage III) Face Region-Focal Alignment, which strengthens the model’s ability to attend to spatially defined facial regions; and (Stage IV) Region-Focal Guided Multi-attribute Recognition, which integrates historical ROI explainable information to perform final multi-attribute decision. This progressive design endows the model with spatial awareness, semantic precision, and interpretable decision-making in localized facial analysis.

\noindent \textbf{Network Architecture.} Focal-RegionFace is built on the Qwen2.5-VL architecture. We use multi-stage LoRA fine-tuning \cite{zheng2024llamafactory} to optimize the base model with face region-focal visual and language reasoning abilities. Initially, each image is processed by Qwen's vision encoder, followed by a learnable projection into the LLM's token embedding space. LoRA modules are applied to critical attention layers, enhancing region-specific representation and multi-attribute reasoning. This structure empowers the model to effectively capture localized facial dynamics and perform fine-grained analysis.

\begin{table*}[ht]
    \centering
    \small
    \vspace{-1em}
    \renewcommand{\arraystretch}{0.8} 
    \setlength{\tabcolsep}{5pt} 
    \resizebox{1\textwidth}{!}{
    \begin{tabular}{@{}l|c|c@{}}
    \toprule
    \textbf{Name} & \textbf{Description} & \textbf{Range} \\ \midrule
    Cls & Matching evaluation of facial detail description and attribute classification. & 0–100 \\
    Det & Descriptive Facial Detail — Richness evolution of facial detail description.  & 0–100 \\
    Flu & Fluency and coherence of the generated language description. & 0–100 \\
    Box & Relevance between regional descriptions and target regions (boxes). & 0–100 \\
    Sem & Semantic alignment of generated descriptions with visual content. & 0–100 \\ 
    \bottomrule
    Win\% & Ratio of samples where the model achieved the highest score. & 0–100 \\
    \bottomrule
    \end{tabular}
    }
    \vspace{-1em}
    \caption{MLLM-based evaluation metric descriptions and corresponding score ranges.}
    \label{tab:Metrics}
\vspace{-1.3em}
\end{table*}

\subsection{Training Strategies} 
In our experiments, we found that single-stage fine-tuning lacks the semantic learning order from perception to understanding to expression. This causes a disconnect between region-level attribute learning and language generation, reducing fine-grained interpretability and consistency. Therefore, we propose a novel multi-stage fine-tuning strategy to address this limitation. In all training stages, the Qwen2.5-VL backbone remains fully frozen, with fine-tuning applied exclusively to the LoRA and projection layers.

\textbf{Stage I: Global-aware Face Perception.}
In the stage I, the model utilizes preprocessed images that without bounding boxes to predict basic facial attributes such as Action Units (AUs), emotions, and age ranges based on global-facial cues. The input query is designed to extract global information. The output is structured as simple labels, e.g. AU3, AU4, Anger, 30–34. To enhance generalization and robustness, we construct five distinct query prompts for each facial attribute thoughout different stages, and randomly assign them to each image. This diverse-prompt strategy improves the model's adaptability across various facial contexts (detailed in the Appendices). This stage establishes the general perception of facial features, enabling the model to have a comprehensive understanding of facial attributes before focusing on specific regions.

\textbf{Stage II: Region-aware face visual-language alignment.} In the stage II, region-specific visual-language alignment is introduced. The input comprises preprocessed images augmented with randomly generated bounding boxes. At this stage, queries are localized, guiding the model to attend exclusively to the visual content within each bounding box. Supervised fine-tuning is performed using detailed natural language descriptions of facial attributes. This process instills the model with an initial understanding of localized regions and their linguistic associations, laying the foundation for more precise localization tasks in subsequent stages.

\textbf{Stage III: Face Region-Focal Alignment.}
To further enhance regional focus, the stage III introduces a Region of Interest (ROI) fine-tuning strategy. The images in Stage III are masked such that only the targeted regions remain in model's interests, while the masked areas are converted to grayscale. This deliberate masking forces the model to generate descriptions exclusively based on aimed content, neglecting global context. The training retains the same structured queries and captions as Stage II. This stage improves model's ability to capture localized expressions, fine lines, and subtle muscular shifts.

\textbf{Stage IV: Region-Focal Guided Multi-attribute Recognition.}
In the final stage, Region-Focal Guided Multi-attribute Recognition emphasizes multi-region aggregation and holistic assessment. The input consists of a single preprocessed facial image annotated with multiple boxed regions, corresponding to the regions defined in Stages II and III. For each region, the model utilizes the fine-grained captions learned previously to perform multi-region reasoning. The results are formatted in the simple ground truth structure from Stage I (e.g., AU3, Anger, 30–34). This stage serves two main purposes: first, to validate the model's capability to integrate detailed observations across multiple regions, and second, to simulate real-world applications where multiple facial areas are queried simultaneously for a unified interpretation. This step finalizes the model's capacity for multi-attribute reasoning across both localized and comprehensive contexts.

\section{Experiment}

\subsection{Experimental Settings}
 \noindent \textbf{Implemental Details.} In each stage, we fine-tune 4-bit quantised Qwen2.5-VL-32B with a batch size of 16, a learning rate of 2e-5, and a cosine learning rate scheduler over 10 epochs. Gradient checkpointing is enabled to reduce memory consumption, and a weight decay of 0.01 is applied for regularization. Further details are provided in the Appendices.

 \noindent \textbf{Evaluation Metrics.} 

We adopt three categories of metrics to comprehensively evaluate Focal-RegionFace.
(1) MLLM-based evaluation metrics (Table \ref{tab:Metrics}) are specifically designed for the new \textit{FaceFocalDesc} task. Leveraging the multimodal reasoning ability of both open- and closed-source MLLMs, we let them act as reviewers to score the generated region-focal descriptions across multiple aspects. This provides an objective and bias-resistant measurement of model reasoning and generation quality.
(2) Mainstream NLP metrics, including BERTScore \cite{zhang2019bertscore} (Precision, Recall, F1), Grammar Issues \cite{languagetoolpython} (GI), and Expert Rating (ER). Thirty experienced annotators, organized into six teams, rated caption quality and semantic alignment, and their scores were aggregated for consensus \cite{cheng2024emotion}.
(3) Traditional recognition metrics, including AU F1 and accuracy for emotion and age prediction.
\begin{table*}[t]
    \centering
    \small
    \vspace{-1em}
    \renewcommand{\arraystretch}{1.3}
    \setlength{\tabcolsep}{3.2pt}

    \begin{adjustbox}{max width=\textwidth}
    \fontsize{7.5}{8.3}\selectfont

    \begin{tabular}{c|ccccccc|ccccccc}
        \toprule
        & \multicolumn{7}{c|}{\cellcolor{gray!15}\textbf{Gemini-2.5-Pro}}
        & \multicolumn{7}{c}{\cellcolor{gray!15}\textbf{GPT-4o}} \\
        \cmidrule(lr){2-8} \cmidrule(lr){9-15}

        \textbf{Model}
        & Cls & Det & Flu & Box & Sem & Win/\% & Rank
        & Cls & Det & Flu & Box & Sem & Win/\% & Rank \\
        \midrule

        Qwen2.5-VL
        & 52.69 & 47.35 & 74.49 & 73.22 & 51.88 & \underline{13.51} & 2
        & 67.28 & \underline{64.62} & \underline{78.20} & \underline{74.73} & \underline{69.68} & 14.57 & 3 \\

        Gemma3
        & \underline{59.04} & \underline{47.69} & 71.40 & 76.53 & \underline{58.01} & 12.40 & 3
        & \underline{67.35} & 60.10 & 72.15 & 71.70 & 68.16 & \underline{19.47} & 2 \\

        Deepseek-Janus-Pro
        & 44.33 & 13.76 & \underline{79.83} & \underline{80.11} & 43.78 & 1.66 & 5
        & 55.20 & 39.91 & 73.41 & 68.19 & 55.98 & 7.55 & 4 \\

        Llama3.2-Vision
        & 51.01 & 33.18 & 74.33 & 68.70 & 45.96 & 4.87 & 4
        & 63.61 & 51.16 & 60.78 & 67.85 & 61.53 & 3.70 & 5 \\

        \midrule
        \textbf{Focal-RegionFace (Ours)}
        & \textbf{70.46} & \textbf{82.91} & \textbf{93.83} & \textbf{91.81} & \textbf{74.70} & \textbf{67.56} & 1
        & \textbf{74.38} & \textbf{83.86} & \textbf{84.72} & \textbf{81.33} & \textbf{79.51} & \textbf{57.71} & 1 \\

        \bottomrule
    \end{tabular}
    \end{adjustbox}

    \vspace{-0.5em}
    \caption{Comparisons of different MLLMs with Focal-RegionFace evaluated by closed-source models.}
    \label{tab:opensource_mllm_eval}
    \vspace{-0.1em}
\end{table*}

\begin{table*}[t]
    \centering
    \small
    \vspace{-1em}
    \renewcommand{\arraystretch}{1.0}
    \fontsize{7.2}{8.0}\selectfont
    \renewcommand\tabcolsep{4.3pt}
    \begin{adjustbox}{max width=1\textwidth}
    \begin{tabular}{c|cccccc}
        \toprule
        & \multicolumn{6}{c}{\cellcolor{gray!15}\textbf{Qwen2.5-VL}} \\
        \cmidrule(lr){2-7}
        \textbf{Comparison} & Cls & Det & Flu & Box & Sem & Win/\% \\
        \midrule
        \begin{tabular}[c]{@{}c@{}} Qwen2.5-VL \\ Focal-RegionFace \end{tabular} &
        \begin{tabular}[c]{@{}c@{}} 65.50 \\ \textbf{75.08} \end{tabular} &
        \begin{tabular}[c]{@{}c@{}} 73.12 \\ \textbf{81.32} \end{tabular} &
        \begin{tabular}[c]{@{}c@{}} 56.45 \\ \textbf{89.67} \end{tabular} &
        \begin{tabular}[c]{@{}c@{}} 75.45 \\ \textbf{81.63} \end{tabular} &
        \begin{tabular}[c]{@{}c@{}} 80.36 \\ \textbf{91.04} \end{tabular} &
        \begin{tabular}[c]{@{}c@{}} \textbf{64.68} \\ 35.32 \end{tabular} \\
        \midrule
        \begin{tabular}[c]{@{}c@{}} Deepseek-Janus-Pro \\ Focal-RegionFace \end{tabular} &
        \begin{tabular}[c]{@{}c@{}} 47.80 \\ \textbf{78.34} \end{tabular} &
        \begin{tabular}[c]{@{}c@{}} 56.99 \\ \textbf{85.14} \end{tabular} &
        \begin{tabular}[c]{@{}c@{}} 35.74 \\ \textbf{92.78} \end{tabular} &
        \begin{tabular}[c]{@{}c@{}} 70.34 \\ \textbf{83.50} \end{tabular} &
        \begin{tabular}[c]{@{}c@{}} 81.35 \\ \textbf{98.13} \end{tabular} &
        \begin{tabular}[c]{@{}c@{}} 14.99 \\ \textbf{85.01} \end{tabular} \\
        \midrule
        \begin{tabular}[c]{@{}c@{}} Llama3.2-Vision \\ Focal-RegionFace \end{tabular} &
        \begin{tabular}[c]{@{}c@{}} 58.90 \\ \textbf{77.06} \end{tabular} &
        \begin{tabular}[c]{@{}c@{}} 61.19 \\ \textbf{83.98} \end{tabular} &
        \begin{tabular}[c]{@{}c@{}} 40.58 \\ \textbf{89.40} \end{tabular} &
        \begin{tabular}[c]{@{}c@{}} 70.68 \\ \textbf{84.26} \end{tabular} &
        \begin{tabular}[c]{@{}c@{}} 81.50 \\ \textbf{94.82} \end{tabular} &
        \begin{tabular}[c]{@{}c@{}} 22.05 \\ \textbf{77.95} \end{tabular} \\

        \midrule\midrule
        & \multicolumn{6}{c}{\cellcolor{gray!15}\textbf{Deepseek-Janus-Pro}} \\
        \cmidrule(lr){2-7}
        \textbf{Comparison} & Cls & Det & Flu & Box & Sem & Win/\% \\
        \midrule
        \begin{tabular}[c]{@{}c@{}} Qwen2.5-VL \\ Focal-RegionFace \end{tabular} &
        \begin{tabular}[c]{@{}c@{}} 76.30 \\ \textbf{89.55} \end{tabular} &
        \begin{tabular}[c]{@{}c@{}} 78.05 \\ \textbf{89.54} \end{tabular} &
        \begin{tabular}[c]{@{}c@{}} 79.71 \\ \textbf{89.51} \end{tabular} &
        \begin{tabular}[c]{@{}c@{}} 78.42 \\ \textbf{88.33} \end{tabular} &
        \begin{tabular}[c]{@{}c@{}} 75.40 \\ \textbf{87.37} \end{tabular} &
        \begin{tabular}[c]{@{}c@{}} 0.00 \\ \textbf{100.00} \end{tabular} \\
        \midrule
        \begin{tabular}[c]{@{}c@{}} Deepseek-Janus-Pro \\ Focal-RegionFace \end{tabular} &
        \begin{tabular}[c]{@{}c@{}} 70.81 \\ \textbf{89.87} \end{tabular} &
        \begin{tabular}[c]{@{}c@{}} 71.96 \\ \textbf{89.96} \end{tabular} &
        \begin{tabular}[c]{@{}c@{}} 73.79 \\ \textbf{89.99} \end{tabular} &
        \begin{tabular}[c]{@{}c@{}} 72.88 \\ \textbf{89.38} \end{tabular} &
        \begin{tabular}[c]{@{}c@{}} 71.38 \\ \textbf{88.48} \end{tabular} &
        \begin{tabular}[c]{@{}c@{}} 0.00 \\ \textbf{100.00} \end{tabular} \\
        \midrule
        \begin{tabular}[c]{@{}c@{}} Llama3.2-Vision \\ Focal-RegionFace \end{tabular} &
        \begin{tabular}[c]{@{}c@{}} 70.09 \\ \textbf{89.61} \end{tabular} &
        \begin{tabular}[c]{@{}c@{}} 71.23 \\ \textbf{89.87} \end{tabular} &
        \begin{tabular}[c]{@{}c@{}} 77.63 \\ \textbf{89.89} \end{tabular} &
        \begin{tabular}[c]{@{}c@{}} 77.61 \\ \textbf{88.14} \end{tabular} &
        \begin{tabular}[c]{@{}c@{}} 77.67 \\ \textbf{87.66} \end{tabular} &
        \begin{tabular}[c]{@{}c@{}} 0.00 \\ \textbf{100.00} \end{tabular} \\

        \midrule\midrule
        & \multicolumn{6}{c}{\cellcolor{gray!15}\textbf{Llama3.2-Vision}} \\
        \cmidrule(lr){2-7}
        \textbf{Comparison} & Cls & Det & Flu & Box & Sem & Win/\% \\
        \midrule
        \begin{tabular}[c]{@{}c@{}} Qwen2.5-VL \\ Focal-RegionFace \end{tabular} &
        \begin{tabular}[c]{@{}c@{}} 59.95 \\ \textbf{80.29} \end{tabular} &
        \begin{tabular}[c]{@{}c@{}} 67.90 \\ \textbf{82.72} \end{tabular} &
        \begin{tabular}[c]{@{}c@{}} 54.88 \\ \textbf{73.29} \end{tabular} &
        \begin{tabular}[c]{@{}c@{}} 70.65 \\ \textbf{82.38} \end{tabular} &
        \begin{tabular}[c]{@{}c@{}} 54.51 \\ \textbf{76.12} \end{tabular} &
        \begin{tabular}[c]{@{}c@{}} 15.46 \\ \textbf{84.54} \end{tabular} \\
        \midrule
        \begin{tabular}[c]{@{}c@{}} Deepseek-Janus-Pro \\ Focal-RegionFace \end{tabular} &
        \begin{tabular}[c]{@{}c@{}} 44.83 \\ \textbf{83.48} \end{tabular} &
        \begin{tabular}[c]{@{}c@{}} 52.85 \\ \textbf{80.25} \end{tabular} &
        \begin{tabular}[c]{@{}c@{}} 43.68 \\ \textbf{75.38} \end{tabular} &
        \begin{tabular}[c]{@{}c@{}} 54.73 \\ \textbf{83.91} \end{tabular} &
        \begin{tabular}[c]{@{}c@{}} 41.75 \\ \textbf{80.42} \end{tabular} &
        \begin{tabular}[c]{@{}c@{}} 7.97 \\ \textbf{92.03} \end{tabular} \\
        \midrule
        \begin{tabular}[c]{@{}c@{}} Llama3.2-Vision \\ Focal-RegionFace \end{tabular} &
        \begin{tabular}[c]{@{}c@{}} 62.97 \\ \textbf{82.77} \end{tabular} &
        \begin{tabular}[c]{@{}c@{}} 68.25 \\ \textbf{87.42} \end{tabular} &
        \begin{tabular}[c]{@{}c@{}} 59.13 \\ \textbf{79.03} \end{tabular} &
        \begin{tabular}[c]{@{}c@{}} 71.74 \\ \textbf{86.69} \end{tabular} &
        \begin{tabular}[c]{@{}c@{}} 64.84 \\ \textbf{81.29} \end{tabular} &
        \begin{tabular}[c]{@{}c@{}} 13.87 \\ \textbf{86.13} \end{tabular} \\
        \bottomrule
    \end{tabular}
    \end{adjustbox}

    \vspace{-0.4em}
    \caption{Comparisons of di   erent MLLMs with Focal-RegionFace by open-source MLLM evaluators.}
    \label{tab:closesource_mllm_eval}
    \vspace{-0.9em}
\end{table*}

\noindent \textbf{MLLM-Based Evaluation Details.} 
To evaluate fine-grained language quality, regional specificity, and semantic alignment, we adopt separate strategies for closed- and open-source models.
For closed-source evaluation, Gemini-2.5-Pro\footnote{https://deepmind.google/technologies/gemini/pro/}
 and GPT-4o\footnote{https://openai.com/index/hello-gpt-4o/}
 act as judges, jointly scoring captions from five models—Focal-RegionFace and four baselines: Llama3.2-Vision, Qwen2.5-VL, Deepseek-Janus-Pro, and Gemma3 \cite{gemmateam2025}. Both judges assess all five captions simultaneously under a unified image-conditioned evaluation prompt designed for fairness.
For open-source evaluation, Llama3.2-Vision, Qwen2.5-VL, and Deepseek-Janus-Pro perform independent one-to-one comparisons between Focal-RegionFace and each baseline using the same evaluation prompt.
This dual strategy ensures fair, standardized, and reproducible assessment across both settings. Prompt details are provided in the Appendices.

\subsection{Experimental Results}

\textbf{I. Quantitative Comparison by the MLLM-based Evaluation.}
To evaluate the effectiveness of our multi-stage training strategy (Figure \ref{fig:framework}), we compare the performance of Focal-RegionFace using both closed-source and open-source MLLMs as intelligent expert evaluators. Due to budget constraints, we adopt global ranking for closed-source models (Table \ref{tab:opensource_mllm_eval}), whereas one-on-one evaluations are conducted for open-source models (Table \ref{tab:closesource_mllm_eval}). 

\begin{table}[t]
    \centering
    \setlength{\tabcolsep}{4.3pt} 
    \fontsize{7.5}{8.3}\selectfont
    \renewcommand{\arraystretch}{1.3} 
    \begin{adjustbox}{max width=1\textwidth}
    \begin{tabular}{l|ccccc}
        \toprule
        \textbf{Model} & \textbf{BS-P} & \textbf{BS-R} & \textbf{BS-F1} & \textbf{GI} ($\downarrow$) & \textbf{ER} \\
        \midrule
        Deepseek-Janus-Pro   & \underline{57.45} & 46.65 & 51.16 & 0.7802 & 34.84  \\
        Llama3.2-Vision      & 53.63 & 52.57 & 52.09 & 2.9200 & 55.23 \\
        Gemma3               & 51.46 & 53.95 & 52.53 & 1.9133 & \underline{78.50}  \\
        Qwen2.5-VL       & 51.67 & \underline{58.62} & \underline{54.84} & 1.6333 & 76.38  \\
        \midrule
        Focal-RegionFace (Ours)    & \textbf{75.55} & \textbf{75.76} & \textbf{75.98} & \textbf{0.4318} & \textbf{86.72}  \\
        \bottomrule
    \end{tabular}
    \end{adjustbox}
    \vspace{-0.8em}
    \caption{Quantitative evaluation of caption quality on NLP metrics, i.e. BERTScore (\%) and Grammar Issues ($\downarrow$ better).}
    \label{tab:NLP metrics}
\end{table}



\begin{table}[t]
    \centering
    \small
    \setlength{\tabcolsep}{4.3pt}
    \fontsize{7.3}{8.3}\selectfont
    \renewcommand{\arraystretch}{1.3}
    \begin{adjustbox}{max width=1\textwidth}
    \begin{tabular}{l|ccc|ccc}
        \toprule
        \multirow{2}{*}{\textbf{Model}} 
        & \multicolumn{3}{c|}{\textbf{Region-Focal}} 
        & \multicolumn{3}{c}{\textbf{Full Face}} \\
        \cmidrule(lr){2-4} \cmidrule(lr){5-7}
        & \textbf{Emo} & \textbf{Age} & \textbf{AU} 
        & \textbf{Emo} & \textbf{Age} & \textbf{AU} \\
        \midrule
        Deepseek-Janus-Pro   
        & 35.21 & 31.92 & 9.21  
        & 41.20 & 36.43 & 14.26  \\
        
        Llama3.2-Vision      
        & 18.42 & 25.18 & 11.56  
        & 38.48 & 37.46 & 18.43  \\
        
        Gemma3     
        & \underline{37.77} & \underline{38.88} & \underline{21.31}  
        & \underline{45.86} & \underline{50.14} & \underline{32.61}  \\
        
        Qwen2.5-VL     
        & 35.64 & 38.11 & 10.06  
        & 45.73 & 47.84 & 24.16  \\
        \midrule
        Focal-RegionFace (Ours)  
        & \textbf{40.35} & \textbf{43.65} & \textbf{23.12}  
        & \textbf{53.74} & \textbf{64.37} & \textbf{40.22}  \\
        \bottomrule
    \end{tabular}
    \end{adjustbox}
    \caption{Quantitative evaluation of multiple attribute recognition using face region-focal images vs. full face images.}
    \label{TableCombined}
    \vspace{-2em}
\end{table}

In general, our results consistently demonstrate that the progressively structured fine-tuning strategy significantly enhances multimodal facial understanding, as reflected in consistently superior performance across all evaluation metrics.

Under the closed-source MLLM-based evaluation, our model consistently outperforms competitive baselines. Notably, among all models, Qwen2.5-VL and Gemma3 exhibit the strongest performance, while Deepseek-Janus-Pro and LLaMA3.2-Vision perform relatively poorly, suggesting that they may be less suitable for facial understanding tasks.

For the open-source model evaluation, we conduct one-on-one comparisons between our model and each baseline using the corresponding open-source MLLMs. Our approach generally achieves consistently better results, with only one exception: against Qwen2.5-VL, our model shows a slightly lower win rate. We hypothesize that this may be due to evaluation bias, where models tend to favor their own outputs over those generated by others, as discussed in  \cite{panickssery2024llm}. The average response time for generating a single description is approximately 0.6s, indicating the model’s potential for real-time interactive applications.

\textbf{II. Mainstream NLP-Metric Evaluation.} 
To enhance the completeness of the evaluation, we also incorporate the main NLP metrics to assess caption generation. As shown in Table \ref{tab:NLP metrics}, Focal-RegionFace exhibits stronger performance on all metrics.  This further highlights that the descriptions generated by Focal-RegionFace have better consistency compared to standard annotations and have fewer grammatical errors.

\textbf{III. Traditional multiattribute recognition evaluation.}
Table \ref{TableCombined} shows the comparisons of our model with other pretrained MLLMs by traditional classification evaluations, including the recognition accuracy of prediction of emotion and age, and the F1-Score of action unit recognition. When we consider only selected regions as image inputs (simulating the face occlusion case), our Focal-RegionFace model recognizes them more accurately and with greater robustness than mainstream MLLMs. When focusing on full face information, our method still maintains the best performance in all attribute recognition tasks.


\subsection{Ablation Study}
\begin{figure}[t]
    \centering
    \includegraphics[width=0.7\textwidth]{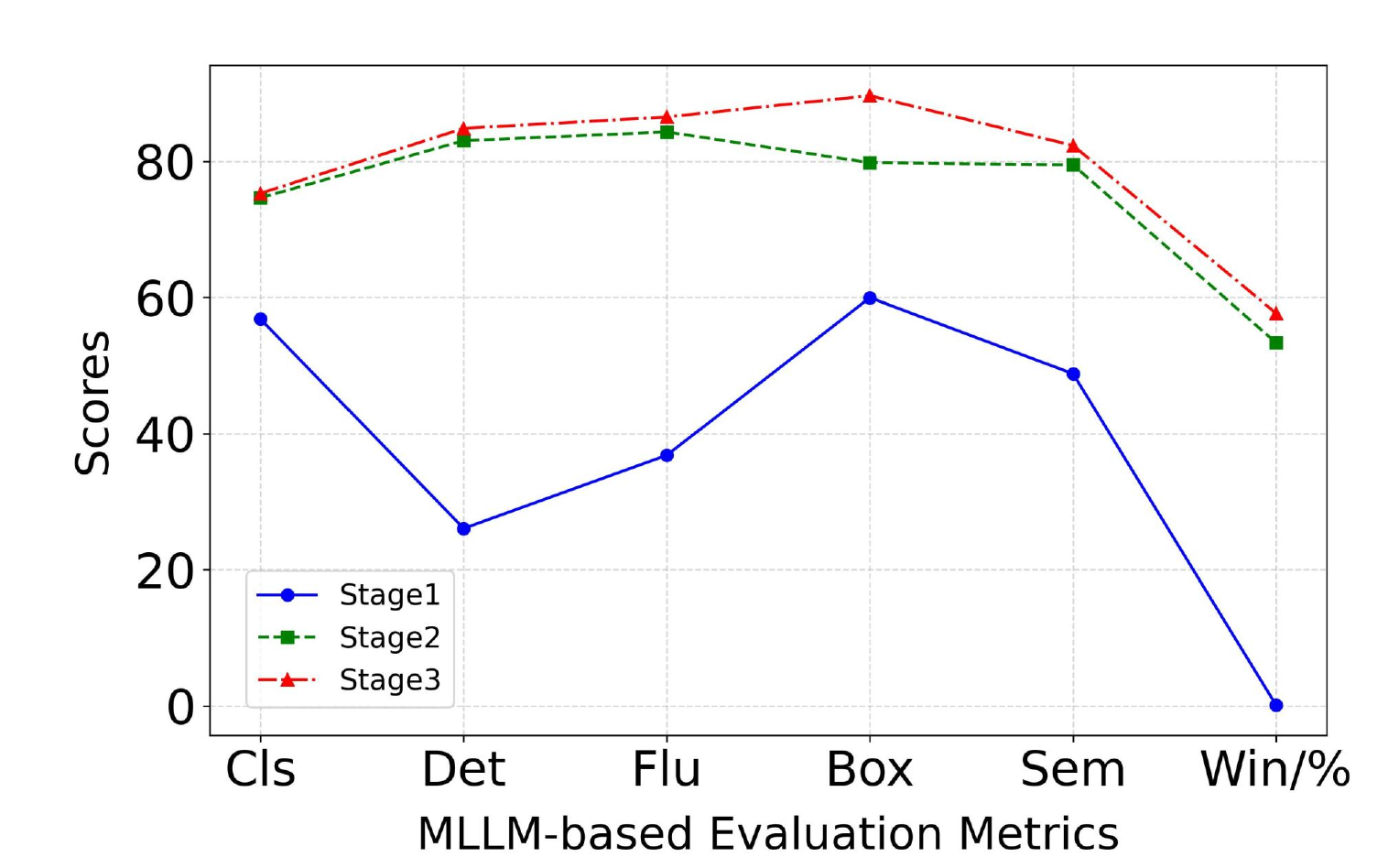}
    \caption{Ablation: MLLM evaluation}
    \label{fig:MLLM_Ablation}
    \vspace{-1.24em}
\end{figure}

\textbf{I. The effect of the multi-stage from I to III:} To understand the impact of each fine-tuning stage in Focal-RegionFace, we perform ablation studies on  where the results are shown in Table \ref{tab:Ablation_Combined} and Figure \ref{fig:MLLM_Ablation}. Compared with the baseline Qwen2.5-VL-32B, in Table \ref{tab:Ablation_Combined}, the performances of multi-attribute recognition are improved by the first stage of face perception fine-tuning. For the multi-attrbute description generations, Figure \ref{fig:MLLM_Ablation} shows that with our multi-stage progressive face region-focal fine-tuning alignments, the multi-attribute descriptions generated by our model achieved significant improvements in several aspects under the closed-source evaluator, i.e. GPT-4o.
In particular, in terms of the scores for the degree of region focusing, our model scores were steadily and significantly improved, from 59.9\% in the first stage, to 79.8\% with the second-stage fine-tuning, and to 89.7\% with the final three-stage region-focal fine-tuning.
In addition, further analysis of the caption quality metrics, as shown in Table \ref{tab:Ablation_Combined}, reveals consistent gains in BS-P, BS-R and BS-F1 (BERTScore) across the three stages. From Stage I to Stage III, the averaged F1 score improves from 31.2\% to 76.0\%, demonstrating enhanced linguistic complexity and fluency as the model's regional awareness deepens.  The GI (Grammar Issues) score is not considered for stage-I, as no sentences are generated at this stage. The GI score after stage-III is lower than the baseline, which demonstrates that multi-stage fine-tuning also improves sentence quality.

These results demonstrate that our progressive fine-tuning enables Focal-RegionFace to capture detailed, region-specific facial attributes more effectively. Detailed breakdowns of each metric are in the Appendices.


\begin{table}[t]
    \centering
    \small
    \label{tab:Ablation_Combined}
    \setlength{\tabcolsep}{4.3pt}
    \fontsize{7.3}{8.3}\selectfont
    \renewcommand{\arraystretch}{1.3}
    \begin{adjustbox}{max width=\textwidth}
    \begin{tabular}{c|ccc|cccc}
        \toprule
        \textbf{Model/Stage} & \textbf{Emo} & \textbf{Age} & \textbf{AU} & \textbf{BS-P} & \textbf{BS-R} & \textbf{BS-F1} & \textbf{GI} \\
        \midrule
        Qwen2.5-VL  & 35.64 & 38.11 & 10.06 & 62.91     & 64.73     & 63.77     & 0.58    \\
        Stage I     & 36.27 & 38.92 & 12.25 & 46.82 & 23.67 & 31.19 & N/A \\
        Stage II    & 37.62     & 38.98     & 12.76     & 72.63 & 71.93 & 72.24 & \textbf{0.31} \\
        Stage III   & \underline{38.33} & \underline{39.35} & \underline{13.17} & \textbf{75.55} & \textbf{75.76} & \textbf{75.98} & 0.43 \\
        Stage IV    & \textbf{40.35} & \textbf{43.65} & \textbf{23.12} & \underline{75.02} & \underline{73.33} & \underline{74.17} & \underline{0.36} \\
        \bottomrule
    \end{tabular}
    \end{adjustbox}
    \vspace{-0.8em}
    \caption{Ablation: multi-attribute and NLP metrics.}
\label{tab:Ablation_Combined}
\end{table}

\textbf{II. The effect of Stage-IV.} To further validate the effectiveness of Stage-IV, we conduct an ablation study under the traditional multi-attribute recognition. As shown in Table \ref{tab:Ablation_Combined}, both Stage III and Stage IV demonstrate significant improvements over the baseline Qwen2.5-VL across all attributes. The introduction of region-focal alignment in Stage-III enhances localized feature extraction, leading to noticeable gains in AU and emotion recognition. In Stage 4, region-focal guided multi-attribute recognition further boosts performance, with AU recognition increasing to 23.12\% and Age prediction reaching 43.65\%, marking a substantial leap compared to previous stages. This progressive refinement confirms the effectiveness of multi-modal multi-region aggregation for fine-grained attribute recognition. However, there is a slight decline in the NLP metrics, which can be attributed to the incorporation of additional classification information that may slightly impact the model’s ability to generate text.

\section{Visualisation} 
\begin{figure*}[t]
  \centering
  \includegraphics[width=1\linewidth]{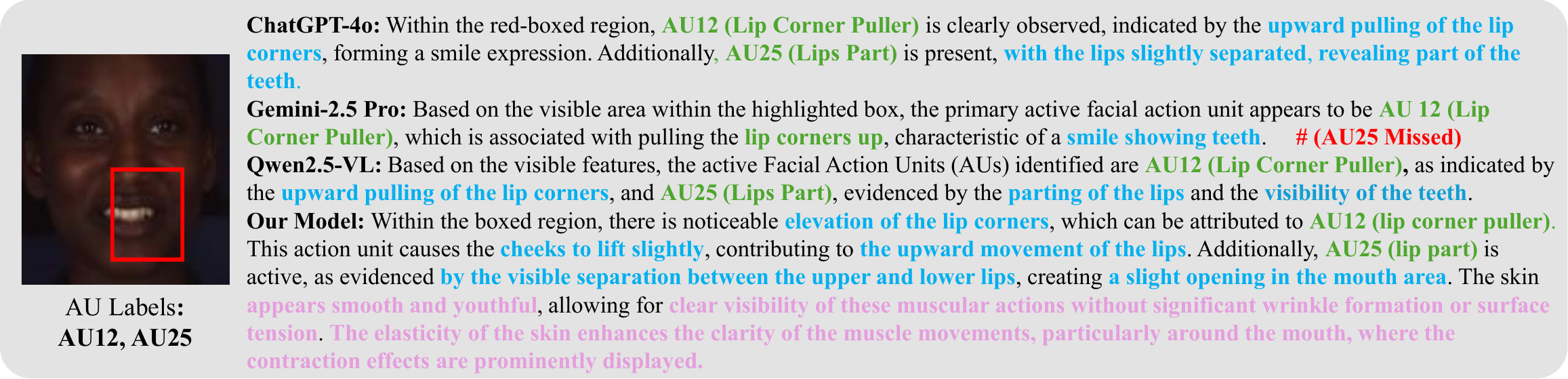}
  \vspace{-0.8em}
  \caption{Visual comparisons of different face state description generators for multiple face attributes, including facial AU, emotion, and age. The red boxes are randomly selected areas. And the descriptions in red are incorrect or region-irrelevant generation. (Blue: AUs description; Green: Muscle description; Purple: Comprehensive analysis of skin details)}
  \label{fig:Visilazation}
  \vspace{-1.2em}
\end{figure*}

To illustrate the effectiveness of Focal-RegionFace on the \textit{FocalDec} task, Figure \ref{fig:Visilazation} presents visual comparisons of generated multi-attribute descriptions across randomly selected regions from multiple subjects, evaluated against ChatGPT-4o, Gemini-2.5-Pro, and Qwen2.5-VL.

Our model excels in localized facial analysis, offering more accurate age estimation through detailed assessment of skin texture, elasticity, and muscle tone, and achieving superior AU detection with precise identification of subtle muscular movements. These physiologically grounded and fine-grained interpretations make predictions both accurate and explainable, demonstrating the model’s strength in region-aware, high-precision facial understanding.

\section{Conclusion}
We introduce \textit{FaceFocalDesc}, a novel task for fine-grained multi-attribute recognition and description generation of arbitrary facial regions, together with MFRF, a benchmark containing 120K region-level annotations and MLLM-based semantic evaluation metrics. To address this task, we propose Focal-RegionFace, a Qwen2.5-VL-based model trained through a four-stage progressive fine-tuning strategy that builds global perception, region-aware alignment, region-focal refinement, and multi-attribute recognition. Experimental results demonstrate that Focal-RegionFace significantly outperforms state-of-the-art MLLMs (e.g., Llama3.2-Vision) in both generation and recognition tasks, achieving superior region-centric facial description performance.

Despite these promising results, several limitations remain. Our study focuses primarily on open-source and closed-source MLLMs under computational constraints; larger-capacity models such as Gemini-2.5-Pro and GPT-4o serve only as evaluators rather than fine-tuning backbones. Additionally, the fine-grained regional annotations in MFRF are generated through a semi-automatic GPT-4o–assisted pipeline, which, despite human refinement, may introduce stylistic inconsistencies or annotation bias. Furthermore, our evaluation relies on judgments from open- and closed-source MLLMs, which can be influenced by model-specific linguistic preferences. Future work may explore scaling Focal-RegionFace to larger models, improving annotation reliability through human–machine collaborative labeling, and developing more robust, cross-model evaluation protocols for fairer and more interpretable assessment.

\appendix


\section{FRFM Dataset Design Method Details}
\subsection{Face Region Selection Method}

\begin{table*}[h!]
\centering
\renewcommand{\arraystretch}{1.0} 
\setlength{\tabcolsep}{10pt} 
\resizebox{1\textwidth}{!}{
\begin{tabular}{|c|c|}
    \hline
    \textbf{Parameter} & \textbf{Description} \\ 
    \hline
    $L$ & Set of facial landmarks, represented as $L = \{(x_i, y_i) | i \in [1, N]\}$ \\ 
    $N$ & Total number of facial landmarks \\ 
    $B_r$ & Set of randomly generated bounding boxes, represented as $B_r = \{(x_1, y_1, x_2, y_2)\}$ \\ 
    $N_b$ & The required number of bounding boxes \\ 
    $IOU_{thresh}$ & Maximum overlap threshold for IoU \\ 
    $W_f, H_f$ & Width and height of the face region \\ 
    $W_{min}, W_{max}$ & Minimum and maximum width of the generated boxes \\ 
    $H_{min}, H_{max}$ & Minimum and maximum height of the generated boxes \\ 
    $M$ & Maximum number of attempts for generating non-overlapping boxes \\ 
    $S_f$ & Final set of successfully generated bounding boxes \\ 
    \hline
\end{tabular}
}
\caption{Details of the parameters used in the face region selection method.}
\label{tab:Face Region Selection Method Parameter Details.}
\end{table*}

Based on the parameters in Table~\ref{tab:Face Region Selection Method Parameter Details.}, we follow the steps below to perform random division of the box regions.

\textbf{Face Region Estimation.}  
Given the set of facial landmarks $L$, the width $W_f$ and height $H_f$ of the face region are computed as:
\begin{multline}
    W_f = \max_{x_i \in L}(x_i) - \min_{x_i \in L}(x_i), \qquad\qquad
    H_f = \max_{y_i \in L}(y_i) - \min_{y_i \in L}(y_i)
\end{multline}
The boundary coordinates of the face region are determined by:
\begin{multline}
(fx_1, fy_1) = (\min_{x_i \in L}(x_i), \min_{y_i \in L}(y_i)), \qquad
(fx_2, fy_2) = (\max_{x_i \in L}(x_i), \max_{y_i \in L}(y_i))
\end{multline}

\textbf{Random Box Generation.}  
The minimum and maximum dimensions for the randomly generated bounding boxes are defined as:
\begin{equation}
    W_{min} = 0.2 \times W_f, \quad W_{max} = 0.4 \times W_f
\end{equation}
\begin{equation}
    H_{min} = 0.2 \times H_f, \quad H_{max} = 0.4 \times H_f
\end{equation}
For each generated bounding box, the coordinates are computed as:
\begin{multline}
x_1 = \text{rand}(fx_1,\, fx_2 - W_{rand}), \qquad\qquad
y_1 = \text{rand}(fy_1,\, fy_2 - H_{rand})
\end{multline}
\begin{equation}
    x_2 = x_1 + W_{rand}, \quad y_2 = y_1 + H_{rand}
\end{equation}
where $W_{rand}$ and $H_{rand}$ are sampled from $[W_{min}, W_{max}]$ and $[H_{min}, H_{max}]$, respectively.

\textbf{Intersection Over Union (IoU).}  
For any two bounding boxes $B_1 = (x_1, y_1, x_2, y_2)$ and $B_2 = (x_3, y_3, x_4, y_4)$:
\begin{equation}
\text{IoU}(B_1, B_2) =
A_{ov} \,/\, (A_1 + A_2 - A_{ov})
\end{equation}
where:
\begin{multline}
\text{Area of Overlap} = 
\max(0, \min(x_2, x_4) - \max(x_1, x_3)) \\
\times \max(0, \min(y_2, y_4) - \max(y_1, y_3))
\end{multline}

\textbf{Iteration Logic.}  
Each time a box is generated, its IoU with all boxes in $S_f$ is checked:
\begin{equation}
    S_f = \{ B_i \;|\; \text{IoU}(B_i, B_j) < IOU_{thresh}, \forall B_j \in S_f \}
\end{equation}
If all IoU values are below $IOU_{thresh}$, the new box is added to $S_f$.

\textbf{Final Generation Process.}
\begin{enumerate}
    \item \textbf{Initialization:} Estimate the face region $W_f, H_f$ from landmarks $L$.
    \item \textbf{Random Sampling:} Generate random bounding boxes up to $M$ attempts:
    \begin{itemize}
        \item Randomly sample coordinates within the facial region.
        \item Compute IoU with existing boxes.
        \item If IoU constraints are satisfied, add the box to $S_f$.
    \end{itemize}
    \item \textbf{Termination:} Repeat until $|S_f| = N_b$.
\end{enumerate}
\vspace{-0.2cm}

\subsection{GPT-4o Generate Prompt Details.}
In this section, we describe the prompt design adopted to ensure that GPT-4o\footnote{https://openai.com/index/hello-gpt-4o/}
 reliably follows our instructions. As stated in the main text, the FRFM prompts are organized into three sequential stages: Contextual Focus, Region Constraint, and Structured Generation.

\begin{figure}[H]
  \centering

  \begin{subfigure}[b]{\linewidth}
    \centering
    \includegraphics[width=0.9\linewidth]{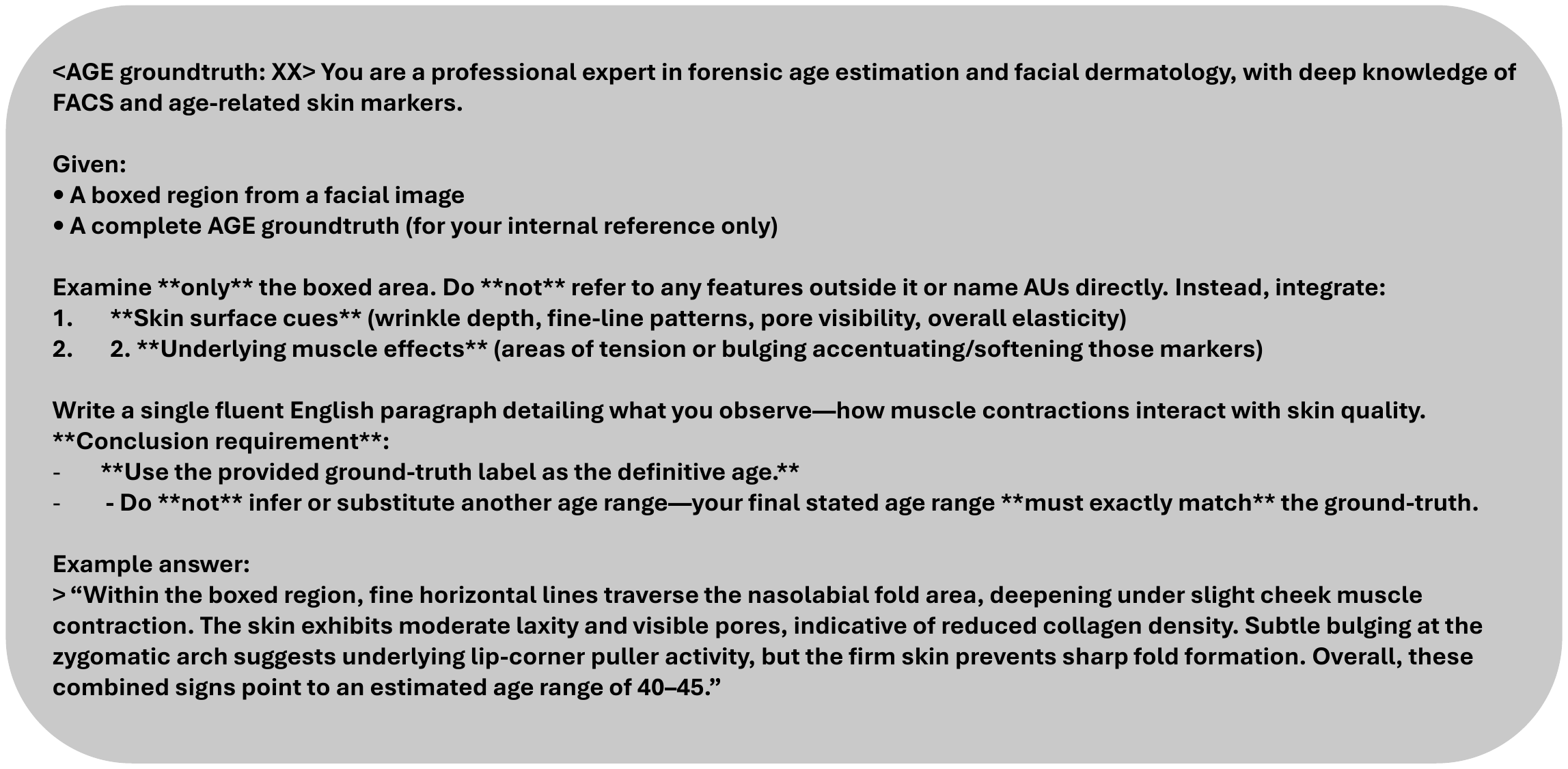}
    \vspace{-0.7em}
    \caption{Prompt details for generating AGE fine-grained descriptions using GPT-4o.}
    \label{fig:AGE GPT-4o Generate Prompt Details}
  \end{subfigure}


  \begin{subfigure}[b]{\linewidth}
    \centering
    \includegraphics[width=0.9\linewidth]{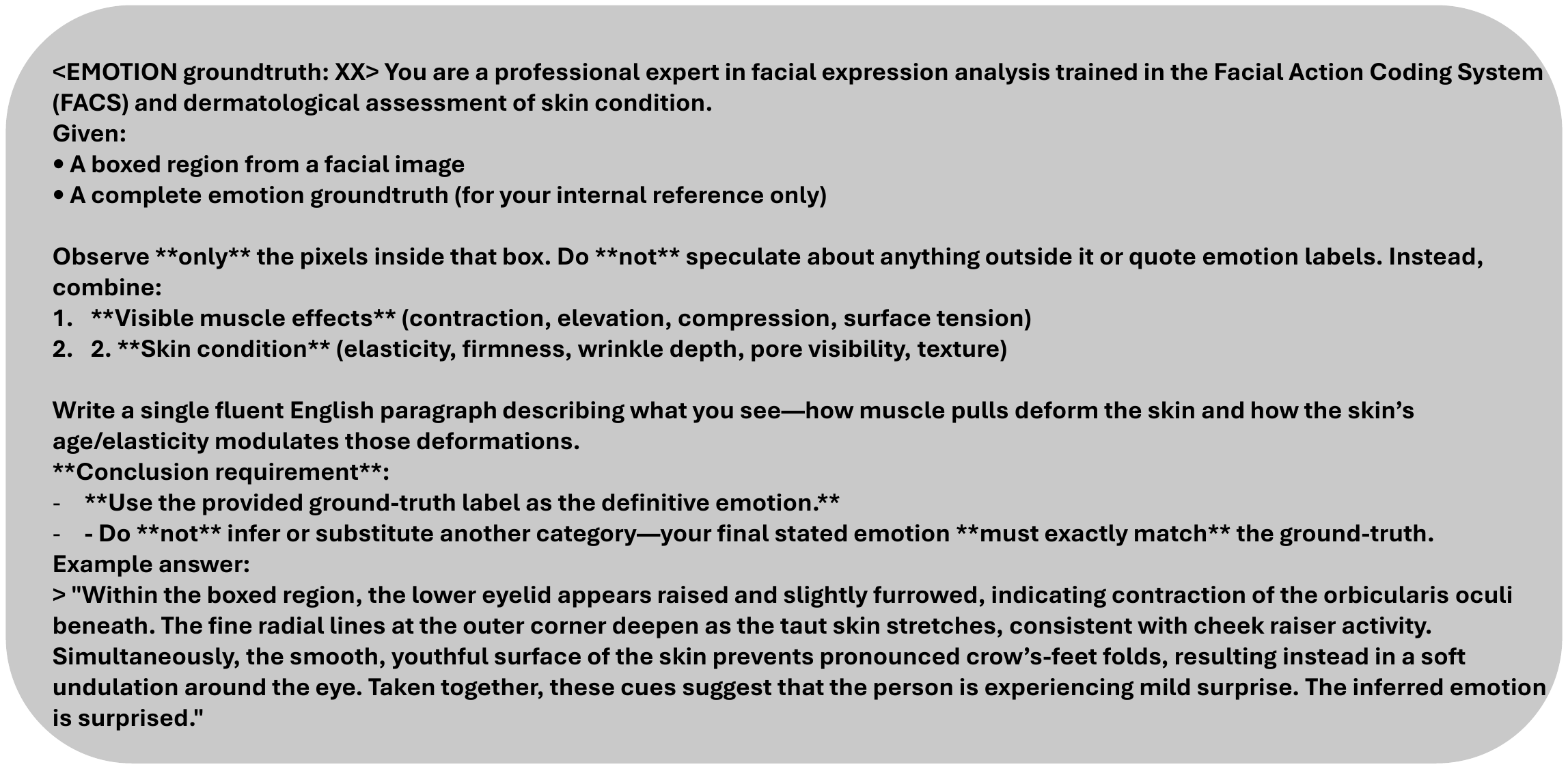}
    \vspace{-0.7em}
    \caption{Prompt details for generating Emotion fine-grained descriptions using GPT-4o.}
    \label{fig:Emotion GPT-4o Generate Prompt Details}
  \end{subfigure}


  \begin{subfigure}[b]{\linewidth}
    \centering
    \includegraphics[width=0.9\linewidth]{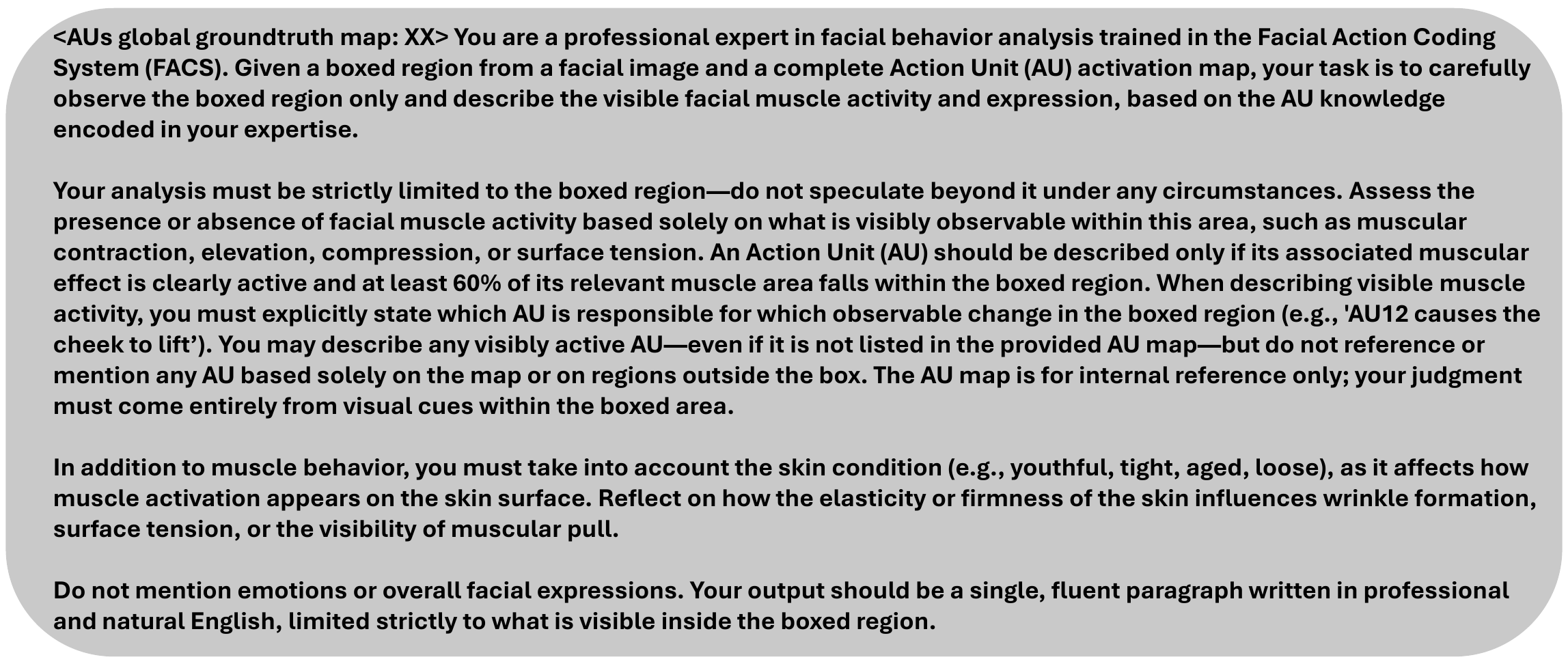}
    \vspace{-0.7em}
    \caption{Prompt details for generating AU fine-grained descriptions using GPT-4o.}
    \label{fig:AU GPT-4o Generate Prompt Details}
  \end{subfigure}

  \caption{Prompt details for generating fine-grained descriptions of AGE, Emotion and AUs using GPT-4o.}
  \label{fig:PromptDesignAll}
\end{figure}

In the Contextual Focus stage, GPT-4o \cite{ray_2023_chatgpt} is assigned an expert role (e.g., a forensic age-estimation and facial dermatology specialist with deep FACS knowledge), as illustrated in Fig.~\ref{fig:AGE GPT-4o Generate Prompt Details}. This role specification anchors the model within the appropriate domain and suppresses irrelevant reasoning. The Region Constraint stage then strictly limits the model’s attention to the boxed facial region (e.g., “Examine only the boxed area”), ensuring that global facial cues do not influence the analysis. Finally, in the Structured Generation stage, GPT-4o is instructed to produce a logically organized paragraph that (1) describes surface-level and muscle-related cues within the box, (2) avoids any out-of-box features or explicit AU references, and (3) concludes with an age estimate consistent with the provided ground-truth label.

For Emotion and Age prompts, GPT-4o is explicitly guided by both the boxed region and the corresponding ground-truth labels. In contrast, AU prompts lack boxed-region AU annotations. To address this, we construct a region-level AU truth map via a two-step process: GPT-4o first selects AUs from the global ground-truth that appear active within the boxed region; it then includes additional AUs if at least 60\% of their canonical activation area falls inside the box. All selected AUs are treated as region-level ground truth. Since boxed-region AU labels are unavailable, the response format for AU prompts remains unconstrained; nevertheless, the analysis is strictly restricted to the specified region and prompt scope, consistent with the Age and Emotion settings.

\section{Training Strategies Details}

\subsection{Diverse-Prompt Strategy (Stage I to Stage IV).} As mentioned in the Stage I section of 3.2, for each Stage (Stage I to Stage IV) of the FRFM dataset, we designed five different queries for each of the three attributes and randomly assigned them to the corresponding attribute images to enhance the model's adaptability to diverse query environments. It is worth noting that we additionally prepend each query with the recognition label of the corresponding attribute: <Task: EMO>, <Task: AU>, or <Task: AGE>.

\begin{figure}[H]
  \centering
  \includegraphics[width=0.7\linewidth]{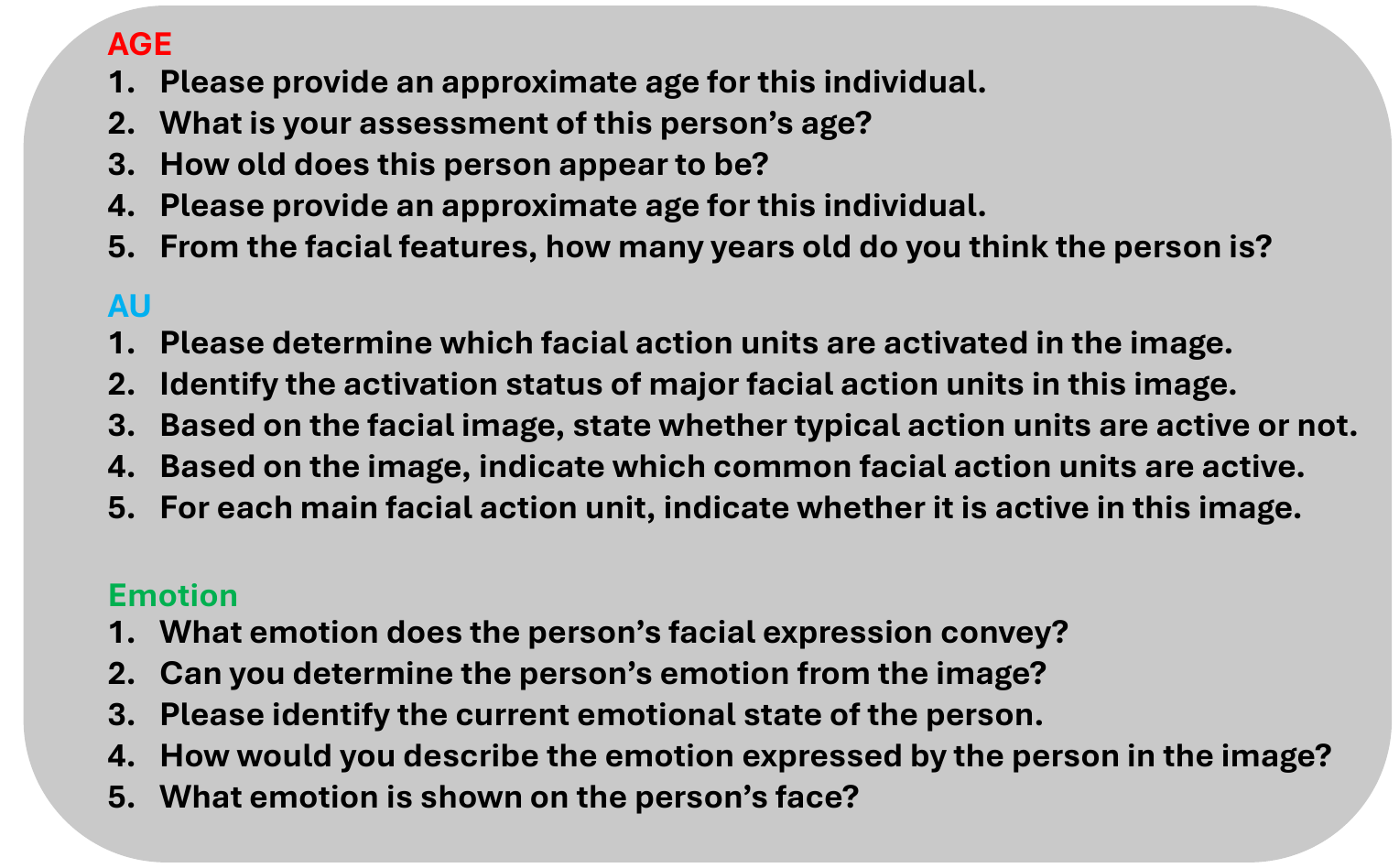}
  \caption{Details of diverse prompts used in Stage I.}
  \label{fig:stage}
\end{figure}
\begin{figure}[H]
  \centering
  \includegraphics[width=0.7\linewidth]{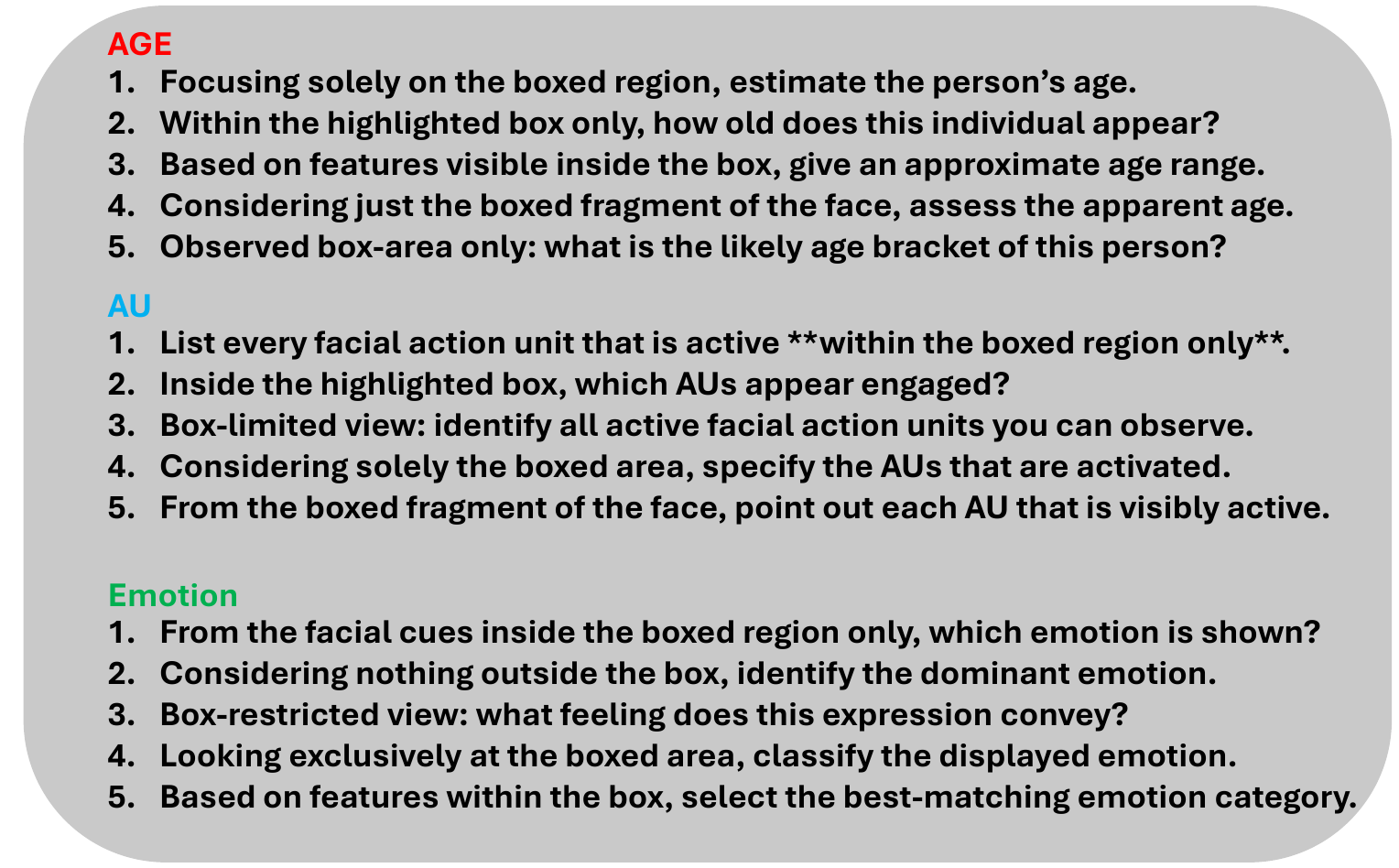}
  \caption{Details of diverse prompts used in Stage II and III.}
  \label{fig:stage2_3}
  \vspace{-1em}
\end{figure}

\begin{figure}[H]
  \centering
  \includegraphics[width=0.7\linewidth]{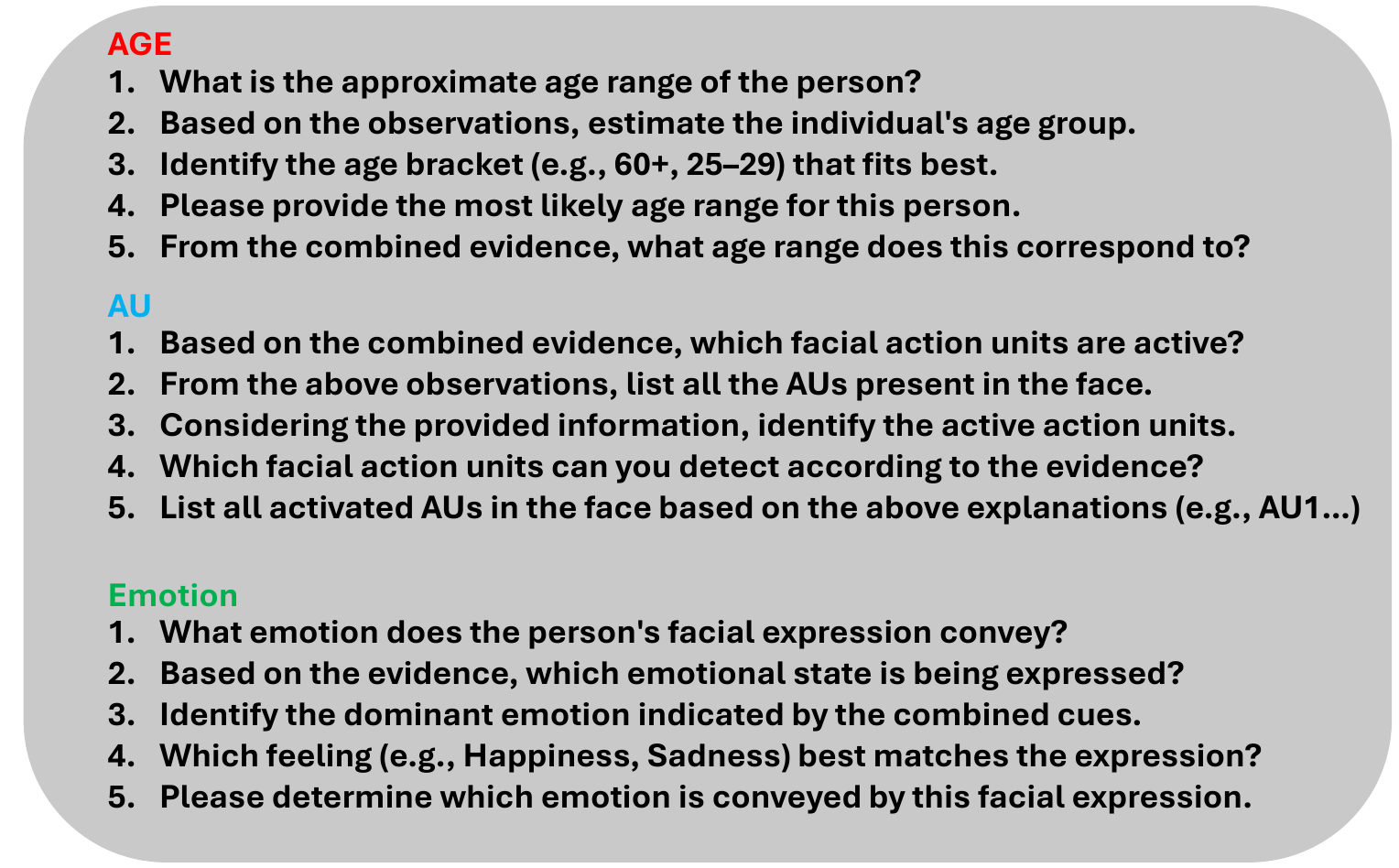}
  \caption{Details of diverse prompts used in Stage IV.}
  \label{fig:stage4}
  \vspace{-1.2em}
\end{figure}

\section{Experimental Details}
\subsection{Implementation Details.} As outlined in Section 4.1 of the main paper, we briefly describe the implementation details of the experimental setup. The baseline model used throughout our experiments is Qwen2.5-32B-VL \cite{Qwen2.5-VL}, with 4-bit quantization applied consistently. As shown in Table~\ref{tab:hyperparams1}, we adopted the same parameter settings across Stage I to Stage III. However, in Stage IV, due to the significant change in input queries, we adjusted the \textit{Cutoff len} while keeping all other parameters unchanged.

\begin{table}[t]
\centering
\caption{Experimental Parameters.}
\renewcommand{\arraystretch}{0.95}
\setlength{\tabcolsep}{10pt}

\begin{tabular}{@{}l c l c@{}}
\toprule
\multicolumn{4}{c}{\textbf{Multi-Stage I--III}} \\ \midrule
\textbf{Parameter} & \textbf{Value} & \textbf{Parameter} & \textbf{Value} \\ \midrule

Training epoch              & 10
& Weight decay               & 0.01 \\

Warmup ratio                & 0.2
& Learning rate              & $2 \times 10^{-5}$ \\

Batch size                  & 16
& Gradient accumulation steps& 4 \\

LR scheduler type            & cosine
& Cutoff length              & 1024 \\

LoRA rank \cite{zheng2024llamafactory}    & 16
& LoRA alpha \cite{zheng2024llamafactory} & 128 \\

LoRA dropout \cite{zheng2024llamafactory} & 0.15
&                                &      \\

\midrule
\multicolumn{4}{c}{\textbf{Focal-RegionFace Stage IV}} \\ \midrule
Cutoff length & 2048 & & \\
\bottomrule
\end{tabular}

\label{tab:hyperparams1}
\end{table}

\subsection{MLLM-Based Evaluation Prompt Setting Details.}
In Section 4.1 outlines the details of our MLLM-Based Evaluation. We carefully designed two types of prompts for evaluating open-source and closed-source models, respectively.

\begin{figure}[H]
  \centering
  \includegraphics[width=0.8\linewidth]{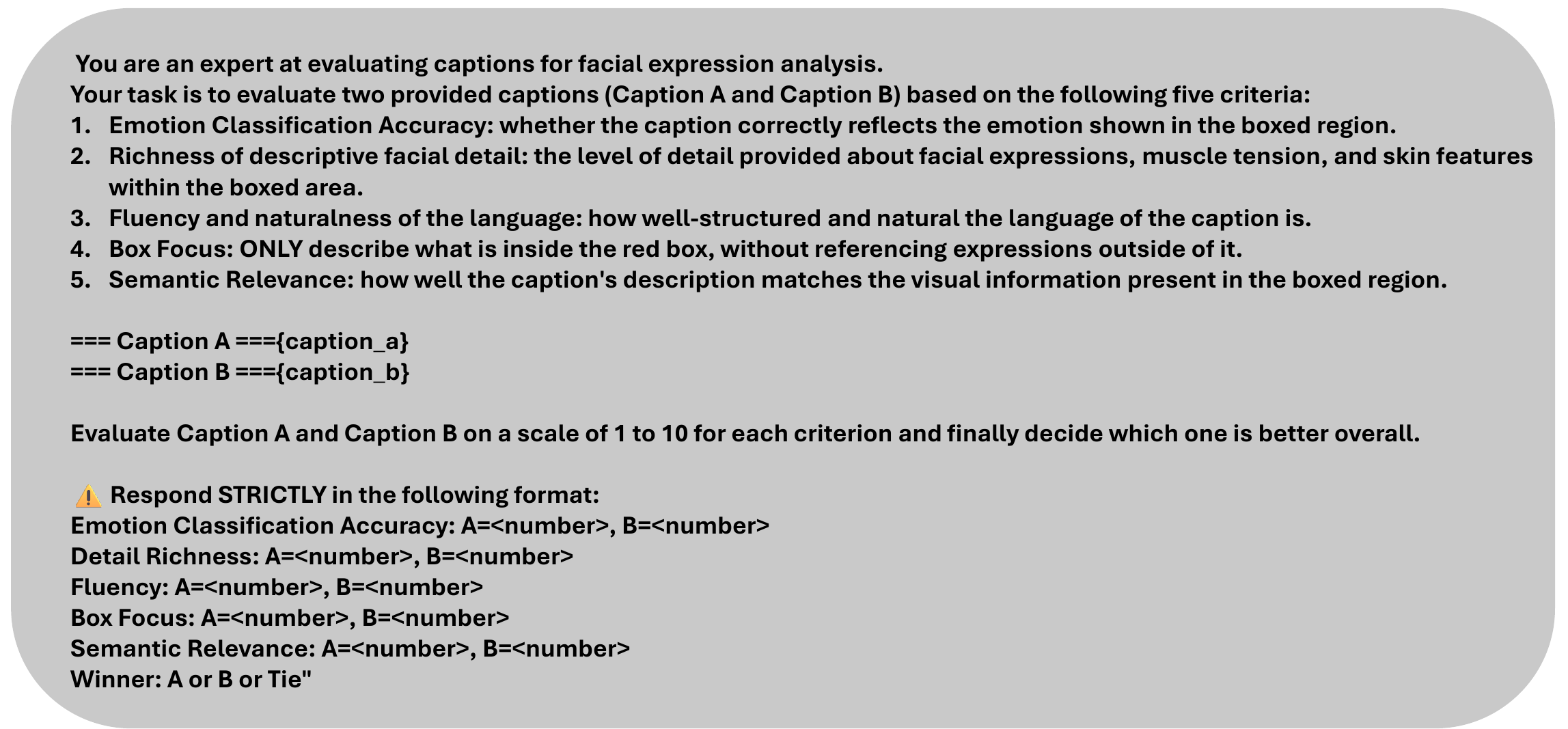}
  \vspace{-0.8em}
  \caption{Details of the open-source evaluation prompts.}
  \label{fig:Opensource_GPT}
  \vspace{-1em}
\end{figure}

As illustrated in Figure~\ref{fig:Opensource_GPT}, the evaluation prompts for open-source models follow a pairwise comparison strategy: for the corresponding image, each evaluation includes two captions, one from our model, another from the other model. This strategy enhances the stability of responses and the accuracy of comparative judgments. In designing the prompt, we first assign a specific role to the LLM, then introduce five key evaluation criteria for the task: \textit{Classification accuracy}, \textit{Richness of descriptive facial detail}, \textit{Fluency and naturalness of the language}, \textit{Box focus}, and \textit{Semantic relevance}. This structure allows for scoring across both task accuracy and caption quality dimensions. Finally, we explicitly define the response format to facilitate downstream parsing and analysis.

\begin{figure}[H]
  \centering
  \includegraphics[width=0.8\linewidth]{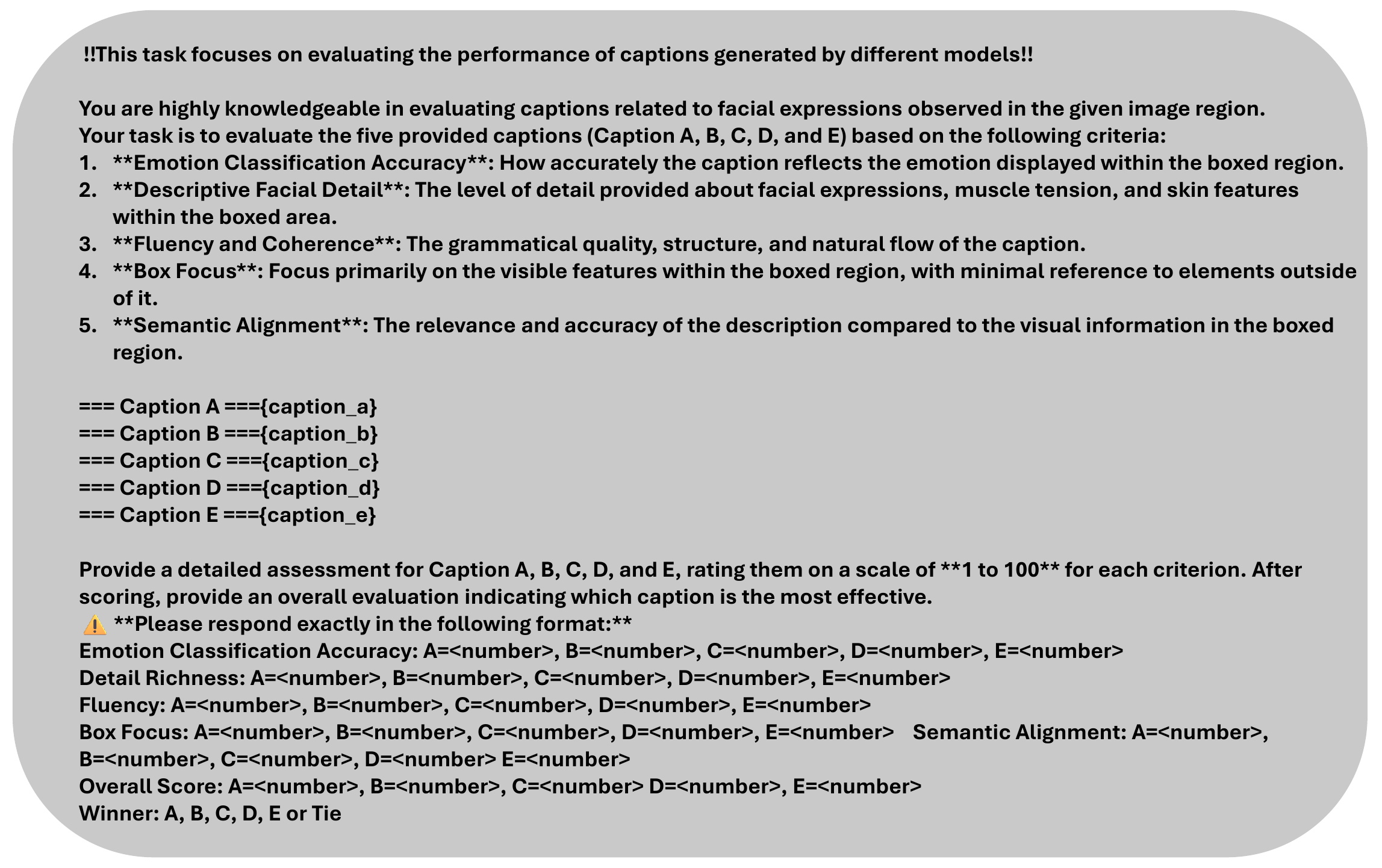}
  \vspace{-1em}
  \caption{Details of the closed-source evaluation prompts.}
  \label{fig:Closedsource_GPT}
  \vspace{-1.8em}
\end{figure}

In contrast, the prompts for closed-source models adopt a multi-caption input strategy, as shown in Figure~\ref{fig:Closedsource_GPT}. This is because closed-source models such as GPT-4o \cite{ray_2023_chatgpt} and Gemini-2.5-Pro \footnote{https://deepmind.google/technologies/gemini/pro/}maintain strong performance and stability even with long contexts and large token inputs. While the structural components of the prompt remain largely the same as in the open-source setup, certain words and sentences were modified to comply with privacy, ethics, and sensitive content constraints imposed by closed-source APIs.

It is important to note that both types of prompts include the corresponding image during inference to support visual-grounded analysis. This design choice leverages the powerful visual-language reasoning capabilities of high-performing models like Gemini-2.5-Pro \cite{geminiteam2025} and ChatGPT-4o \cite{ray_2023_chatgpt}. By jointly inputting the image and multiple captions, we obtain more reliable and fine-grained evaluation outcomes. However, we stress that the results from closed-source evaluations are not intended to replace the experiments with open-source models. Instead, they serve as complementary references to help us achieve a more comprehensive and balanced understanding of caption quality across different model paradigms.
\vspace{-1em}

\subsection{MLLM-based Detailed Metric Breakdowns.}
To complement the main evaluation results, we provide a more fine-grained analysis across multiple facial understanding dimensions—AGE, AU, and EMO—under both open-source (Table~\ref{tab:ap_mllm_eval_open_source}) and closed-source MLLMs (Table~\ref{tab:ap_mllm_eval_close_source}). We report detailed scores for each evaluation criterion (e.g., Cls, Det, Flu, etc.). Our Focal-RegionFace consistently achieves strong performance across nearly all metrics and settings for the three dimensions, further demonstrating the effectiveness of our approach.

Notably, during both open-source and closed-source model evaluations, there are instances where a model scores higher than others on multiple metrics but ends up with a lower overall win rate. This phenomenon is particularly evident when Qwen2.5-VL \cite{Qwen2.5-VL} acts as the evaluator comparing itself with our model. The main reason lies in the presence of outlier scores and fluctuating metric values across certain test samples. Although the final average scores may appear high, the win rate can still be lower. This occurs because high-performing LLMs, when serving as evaluators, often exhibit a bias toward captions generated by themselves. In our experiments, since our model frequently produces captions superior to those of Qwen2.5-VL \cite{Qwen2.5-VL}, this leads to inconsistent scoring—some metrics favor Qwen2.5-VL \cite{Qwen2.5-VL} while others favor ours—resulting in high average scores but fewer individual wins.

\begin{table*}[h!]
    \centering
    \caption{Detailed breakdown of each metric based on closed source MLLM evaluation on multiple attribute predictions (AGE, AU, and Emotion).}
    \renewcommand{\arraystretch}{1.3} 
    \setlength{\tabcolsep}{6pt} 
    \resizebox{0.95\textwidth}{!}{
    \begin{tabular}{c|ccccccc|ccccccc}
        \toprule
        & \multicolumn{7}{c|}{\cellcolor{gray!15}\textbf{Gemini-2.5-Pro} \cite{geminiteam2025}} & 
        \multicolumn{7}{c}{\cellcolor{gray!15}\textbf{GPT-4o} \cite{ray_2023_chatgpt}} \\
        \cmidrule(lr){0-1} \cmidrule(lr){2-8} \cmidrule(lr){9-15}
        \multicolumn{15}{c}{\textbf{AGE}} \\ 
        \midrule
        \textbf{Model} & Cls & Det & Flu & Box & Sem & Win/\% & Rank & Cls & Det & Flu & Box & Sem & Win/\% & Rank \\ 
        \midrule
        \textbf{Qwen2.5-VL \cite{Qwen2.5-VL}} & 70.73 & 62.80 & 88.33 & 79.51 & 66.57 & 14.89 & 3 & 78.20 & 82.86 & \underline{86.93} & 81.18 & 82.68 & 7.32 & 4\\
        \textbf{Gemma3 \cite{gemmateam2025}} & \textbf{81.76} & \underline{66.76} & 88.99 & 80.63 & \underline{78.04} & \underline{15.96} & 2 & \underline{85.83} & \underline{84.24} & 86.67 & \textbf{85.07} & \textbf{91.2} & \underline{39.61} & 2\\
        \textbf{Deepseek-Janus-Pro \cite{chen2025janus}} & 73.90 & 20.33 & \underline{89.65} & \underline{88.00} & 68.58 & 1.04 & 5 & 76.33 & 56.06 & 78.88 & 82.36 & 75.87 & 8.32 & 3\\
        \textbf{Llama3.2-Vision \cite{lee2025efficient}} & 69.97 & 42.33 & 70.97 & 67.85 & 56.71 & 1.71 & 4 & 73.22 & 59.35 & 64.59 & 73.99 & 73.51 & 0.40 & 5\\
        \bottomrule
        \textbf{Focal-RegionFace} & \underline{77.84} & \textbf{87.21} & \textbf{94.50} & \textbf{93.56} & \textbf{78.56} & \textbf{66.40} & 1 & \textbf{86.55} & \textbf{88.06} & \textbf{92.43} & \underline{83.01} & \underline{90.06} & \textbf{52.30} & 1\\
        \midrule
        \multicolumn{15}{c}{\textbf{AU}} \\ 
        \midrule
        \textbf{Qwen2.5-VL \cite{Qwen2.5-VL}} & \underline{33.67} & \underline{38.94} & \underline{72.25} & 68.77 & \underline{37.87} & \underline{9.57} & 2 & \underline{59.33} & \underline{66.56} & \underline{76.03} & \underline{65.31} & \underline{65.62} & \underline{16.65} & 2\\
        \textbf{Gemma3 \cite{gemmateam2025}} & 31.91 & 32.15 & 40.82 & \underline{69.30} & 35.84 & 8.95 & 3 & 49.18 & 34.13 & 47.80 & 51.84 & 49.21 & 3.48 & 5\\
        \textbf{Deepseek-Janus-Pro \cite{chen2025janus}} & 8.08 & 9.11 & 59.55 & 61.67 & 12.18 & 0.75 & 5 & 29.97 & 30.24 & 58.05 & 47.49 & 35.28 & 8.72 & 3\\
        \textbf{Llama3.2-Vision \cite{lee2025efficient}} & 32.02 & 23.45 & 65.77 & 67.48 & 33.89 & 7.73 & 4 & 53.13 & 38.96 & 53.59 & 59.08 & 56.79 & 7.61 & 4\\
        \bottomrule
        \textbf{Focal-RegionFace} & \textbf{67.04} & \textbf{79.25} & \textbf{93.34} & \textbf{89.30} & \textbf{73.23} & \textbf{72.68} & 1 & \textbf{75.94} & \textbf{87.89} & \textbf{88.66} & \textbf{72.88} & \textbf{80.88} & \textbf{71.43} & 1\\
        \midrule
        \multicolumn{15}{c}{\textbf{Emotion}} \\ 
        \midrule
        \textbf{Qwen2.5-VL \cite{Qwen2.5-VL}} & 53.67 & 40.30 & 62.89 & 71.39 & 51.19 & \underline{15.31} & 2 & 64.30 & \underline{44.44} & 71.65 & 77.71 & 60.73 & \underline{28.48} & 2\\
        \textbf{Gemma3 \cite{gemmateam2025}} & \underline{63.45} & \underline{44.16} & 84.40 & 79.66 & \underline{60.14} & 12.28 & 3 & \underline{67.04} & 30.21 & 80.24 & 74.13 & \underline{64.11} & 15.31 & 3\\
        \textbf{Deepseek-Janus-Pro \cite{chen2025janus}} & 51.01 & 11.84 & \underline{90.29} & \underline{90.67} & 50.59 & 3.19 & 5 & 56.23 & 20.42 & 82.41 & \underline{74.71} & 56.80 & 11.61 & 4\\
        \textbf{Llama3.2-Vision \cite{lee2025efficient}} & 51.03 & 33.75 & 86.24 & 70.77 & 47.29 & 5.17 & 4 & 49.48 & 31.13 & \underline{83.87} & 74.47 & 54.28 & 3.1 & 5\\
        \bottomrule
        \textbf{Focal-RegionFace} & \textbf{66.51} & \textbf{82.27} & \textbf{93.65} & \textbf{92.58} & \textbf{72.30} & \textbf{63.61} & 1 & \textbf{71.80} & \textbf{75.62} & \textbf{85.08} & \textbf{88.10} & \textbf{67.58} & \textbf{49.39} & 1\\
        \bottomrule
    \end{tabular}
    }
    \label{tab:ap_mllm_eval_close_source}
\end{table*}

\begin{table*}[h!]
    \centering
    \caption{Detailed breakdown of each metric based on open source MLLM evaluation on multiple attribute predictions (AGE, AU, and Emotion).}
    \renewcommand{\arraystretch}{1.4} 
    \setlength{\tabcolsep}{4pt} 
    \resizebox{0.95\textwidth}{!}{
    \begin{tabular}{c|cccccc|cccccc|cccccc}
        \toprule
        & \multicolumn{6}{c|}{\cellcolor{gray!15}\textbf{Qwen2.5-VL \cite{Qwen2.5-VL}}} & 
        \multicolumn{6}{c|}{\cellcolor{gray!15}\textbf{Deepseek-Janus-Pro \cite{chen2025janus}}} &
        \multicolumn{6}{c}{\cellcolor{gray!15}\textbf{Llama3.2-Vision \cite{lee2025efficient}}} \\
        \cmidrule(lr){0-1} \cmidrule(lr){2-7} \cmidrule(lr){8-13} \cmidrule(lr){14-19}
        \multicolumn{19}{c}{\textbf{AGE}} \\ 
        \midrule
        \textbf{Model} & Cls & Det & Flu & Box & Sem & Win/\% & Cls & Det & Flu & Box & Sem & Win/\% & Cls & Det & Flu & Box & Sem & Win/\% \\ 
        \midrule
        Qwen2.5-VL  \cite{Qwen2.5-VL}     & 68.41 & \underline{80.38} & 55.12 & 76.35 & 80.46 & \textbf{58.34} & 79.49 & 79.90 & 81.44 & 78.92 & 75.24 & 0.07 & 58.09 & 67.77 & 53.25 & 74.49 & 62.27 & 18.91 \\
        Focal-RegionFace & \underline{73.64} & 73.47 & \underline{91.45} & \underline{81.37} & \underline{92.23} & 41.66 & \underline{88.82} & \underline{91.05} & \underline{91.38} & \underline{89.81} & \underline{89.67} & \textbf{99.93} & \underline{77.80} & \underline{84.02} & \underline{70.98} & \underline{80.60} & \underline{72.24} & \textbf{81.09} \\
        \midrule
        Deepseek-Janus-Pro \cite{chen2025janus} & 50.18 & 56.11 & 38.55 & 73.01 & 83.47 & 10.18 & 81.32 & 82.46 & 81.31 & 82.33 & 83.71 & 0.11 & 44.32 & 55.78 & 46.67 & 62.66 & 50.13 & 6.25 \\
        Focal-RegionFace   & \underline{77.44} & \underline{85.59} & \underline{93.12} & \underline{86.39} & \underline{96.53} & \textbf{89.82} & \underline{89.22} & \underline{89.48} & \underline{90.46} & \underline{89.37} & \underline{89.11} & \textbf{99.89} & \underline{83.43} & \underline{75.56} & \underline{65.39} & \underline{80.09} & \underline{71.83} & \textbf{93.75} \\
        \midrule
        Llama3.2-Vision \cite{lee2025efficient}   & 64.41 & 69.82 & 53.78 & 75.37 & 81.60 & 37.04 & 79.90 & 80.98 & 84.18 & 82.64 & 84.97 & 0.05 & 63.17 & 73.06 & 61.94 & 76.30 & 67.31 & 11.96 \\
        Focal-RegionFace   & \underline{77.00} & \underline{83.67} & \underline{87.43} & \underline{82.38} & \underline{88.33} & \textbf{62.96} & \underline{88.23} & \underline{89.04} & \underline{89.90} & \underline{89.03} & \underline{89.12} & \textbf{99.95} & \underline{78.24} & \underline{85.68} & \underline{76.22} & \underline{84.12} & \underline{76.32} & \textbf{88.04} \\
        \midrule
        \multicolumn{19}{c}{\textbf{AU}} \\ 
        \midrule
        Qwen2.5-VL \cite{Qwen2.5-VL}      & 64.20 & 73.99 & 57.13 & 75.20 & 80.72 & \textbf{62.02} & 76.90 & 78.29 & 81.23 & 78.55 & 75.01 & 0.00 & 60.81 & 69.17 & 54.17 & 70.40 & 62.05 & 20.00 \\
        Focal-RegionFace & \underline{76.82} & \underline{81.25} & \underline{79.10} & \underline{80.77} & \underline{92.99} & 37.98 & \underline{87.16} & \underline{89.28} & \underline{89.23} & \underline{86.92} & \underline{86.62} & \textbf{100.00} & \underline{79.70} & \underline{83.43} & \underline{72.12} & \underline{82.32} & \underline{76.34} & \textbf{80.00} \\
        \midrule
        Deepseek-Janus-Pro \cite{chen2025janus} & 48.22 & 57.16 & 36.60 & 71.77 & 79.45 & 12.52 & 77.40 & 79.14 & 81.59 & 80.01 & 79.42 & 0.00 & 43.82 & 53.31 & 43.21 & 58.41 & 47.98 & 9.15 \\
        Focal-RegionFace   & \underline{80.42} & \underline{86.09} & \underline{92.96} & \underline{83.62} & \underline{94.58} & \textbf{87.48} & \underline{91.94} & \underline{91.82} & \underline{92.24} & \underline{91.31} & \underline{92.07} & \textbf{100.00} & \underline{83.22} & \underline{78.39} & \underline{70.77} & \underline{82.52} & \underline{77.81} & \textbf{90.85} \\
        \midrule
        Llama3.2-Vision \cite{lee2025efficient}   & 57.19 & 66.41 & 49.54 & 72.46 & 83.42 & 36.34 & 71.77 & 72.97 & 80.36 & 80.17 & 82.45 & 0.00 & 64.13 & 71.28 & 58.51 & 71.59 & 64.77 & 14.80 \\
        Focal-RegionFace   & \underline{78.48} & \underline{85.36} & \underline{89.53} & \underline{83.90} & \underline{90.76} & \textbf{63.66} & \underline{87.97} & \underline{89.14} & \underline{91.08} & \underline{88.73} & \underline{88.67} & \textbf{100.00} & \underline{81.36} & \underline{88.98} & \underline{77.19} & \underline{84.81} & \underline{76.12} & \textbf{85.20} \\
        \midrule
        \multicolumn{19}{c}{\textbf{Emotion}} \\ 
        \midrule
        Qwen2.5-VL \cite{Qwen2.5-VL}      & 63.89 & 72.90 & 56.70 & 74.12 & 80.00 & 34.98 & 72.51 & 78.57 & 81.09 & 79.41 & 75.84 & 0.00 & 60.95 & 66.91 & 55.93 & 71.85 & 59.21 & 14.26 \\
        Focal-RegionFace & \underline{75.83} & \underline{82.70} & \underline{97.44} & \underline{82.09} & \underline{91.90} & \textbf{65.02} & \underline{92.67} & \underline{92.29} & \underline{91.95} & \underline{91.49} & \underline{92.44} & \textbf{100.00} & \underline{85.45} & \underline{83.63} & \underline{73.99} & \underline{83.67} & \underline{82.51} & \textbf{85.74} \\
        \midrule
        Deepseek-Janus-Pro \cite{chen2025janus} & 45.01 & 56.31 & 26.86 & 73.32 & 82.32 & 8.10 & 70.18 & 80.05 & 81.79 & 79.61 & 80.12 & 0.00 & 46.33 & 53.76 & 36.56 & 56.96 & 44.23 & 3.78 \\
        Focal-RegionFace   & \underline{81.25} & \underline{86.29} & \underline{93.01} & \underline{84.37} & \underline{94.21} & \textbf{91.90} & \underline{92.50} & \underline{92.43} & \underline{92.64} & \underline{91.78} & \underline{92.38} & \textbf{100.00} & \underline{87.25} & \underline{80.74} & \underline{71.22} & \underline{81.90} & \underline{75.35} & \textbf{96.22} \\
        \midrule
        Llama3.2-Vision \cite{lee2025efficient}   & 54.10 & 65.32 & 41.56 & 73.10 & 79.86 & 17.62 & 67.57 & 75.04 & 83.77 & 82.33 & 82.90 & 0.00 & 60.29 & 69.63 & 58.74 & 76.80 & 62.41 & 8.98 \\
        Focal-RegionFace   & \underline{78.85} & \underline{81.13} & \underline{88.47} & \underline{76.90} & \underline{88.77} & \textbf{82.38} & \underline{89.56} & \underline{89.67} & \underline{90.77} & \underline{89.86} & \underline{89.69} & \textbf{100.00} & \underline{76.60} & \underline{85.70} & \underline{74.53} & \underline{81.63} & \underline{74.11} & \textbf{91.02} \\
        \bottomrule
    \end{tabular}
    }
    \label{tab:ap_mllm_eval_open_source}
\end{table*}

\subsection{Traditional Multi-attribute Recognition Evaluation Details.} 
In Section 4.3-III, we presented the average performance of our model across AU, emotion, and age recognition tasks, demonstrating its superiority over most existing open-source LLMs in traditional multi-attribute recognition. In this section, we provide further details to support those results. As shown in Table \ref{tab:Traditional_evaluation_detail}, our model achieves either the best or second-best performance in the majority of classes for each attribute recognition task, indicating its consistent strength across various categories.

\begin{table*}[h!]
\centering
\caption{Detailed breakdown of each metric based on open-source MLLM evaluation on multiple attribute predictions (AGE, AU, and Emotion) using face region-focal images. The Evaluation metric is accuracy and F1-score (\%).}
\label{tab:Traditional_evaluation_detail}
\setlength{\tabcolsep}{5pt} 
\renewcommand{\arraystretch}{1.2} 
\begin{adjustbox}{max width=0.95\textwidth}
\begin{tabular}{l|ccccccc|c}
    \toprule
    \textbf{Model} & \textbf{Neutral} & \textbf{Anger} & \textbf{Disgust} & \textbf{Fear} & \textbf{Happiness} & \textbf{Sadness} & \textbf{Surprise} & \textbf{Avg. (\%)} \\
    \midrule
    Deepseek-Janus-Pro \cite{chen2025janus}  & \underline{4.44} & \textbf{85.28} & 0.00 & 17.86 & 43.10 & 15.48 & 80.34 & 35.21  \\

    Llama3.2-Vision \cite{lee2025efficient}     & 0.00 & 36.61 & 17.86 & 3.57 & 28.16 & 3.57  & \underline{44.64} & 18.42  \\

    Gemma3 \cite{gemmateam2025}             & 1.67   & 50.57    & \underline{18.45}    & 27.38   & \underline{63.09}    & \textbf{78.57}    & 24.69    & \underline{37.77} \\

    Qwen2.5-VL \cite{Qwen2.5-VL}      & \textbf{95.89} & \underline{55.17} & 2.38 & \underline{29.17}  & 26.44 & 16.67 & 23.81 & 35.64  \\
    \bottomrule
    Focal-RegionFace           & 2.22 & 51.15 & \textbf{22.62} & \textbf{32.14} & \textbf{68.97} & \underline{77.38} & 27.98 & \textbf{40.35}  \\

    \bottomrule
\end{tabular}
\end{adjustbox}


\begin{adjustbox}{max width=0.95\textwidth}
\begin{tabular}{l|cccccccccccc|c}
    \toprule
    \textbf{Model} & \textbf{0-4} & \textbf{5-9} & \textbf{10-14} & \textbf{15-19} & \textbf{20-24} & \textbf{25-29} & \textbf{30-34} & \textbf{35-39} & \textbf{40-44} & \textbf{45-49} & \textbf{50-59} & \textbf{60+} & \textbf{Avg. (\%)} \\
    \midrule
    Deepseek-Janus-Pro \cite{chen2025janus}  & 17.95 & 15.04 & \textbf{67.73} & 17.53 & \underline{61.79} & 30.00 & \textbf{46.03} & 0.00 & 15.87 & 0.00 & 18.75 & 75.64 & 31.92 \\

    Llama3.2-Vision \cite{lee2025efficient}  & \textbf{96.83} & 1.63 & 15.87 & 2.78 & \textbf{92.28} & 0.00 & 0.81 & 0.00 & 0.00 & 0.00 & 0.00 & \textbf{93.50} & 25.18 \\

    Gemma3 \cite{gemmateam2025} & 88.10 & 40.65 & \underline{50.00} & 25.79 & 39.43 & \textbf{33.33} & \underline{38.21} & 10.71 & 15.87 & 2.85 & \textbf{59.13} & 62.20 & \underline{38.88} \\

    Qwen2.5-VL \cite{Qwen2.5-VL}  & 88.49 & \textbf{63.82} & 26.59 & \underline{33.73} & 54.07 & \underline{30.95} & 17.07 & \underline{17.46} & \textbf{18.65} & \textbf{14.63} & 10.32 & 82.52 & 38.11 \\
    \bottomrule 
    Focal-RegionFace & \underline{91.46} & \underline{55.28} & 48.41 & \textbf{49.21} & 45.53 & 11.11 & 2.44 & \textbf{59.52} & \underline{18.25} & \underline{12.60} & \underline{39.29} & \underline{90.65} & \textbf{43.65} \\
    \bottomrule
\end{tabular}
\end{adjustbox}


\begin{adjustbox}{max width=0.95\textwidth}
\begin{tabular}{l|cccccccccccc|c}
    \toprule
    \textbf{Model} & \textbf{AU1} & \textbf{AU2} & \textbf{AU4} & \textbf{AU6} & \textbf{AU7} & \textbf{AU10} & \textbf{AU12} & \textbf{AU14} & \textbf{AU15} & \textbf{AU17} & \textbf{AU23} & \textbf{AU24} & \textbf{F1-Score}\\
    \midrule
    Deepseek-Janus-Pro \cite{chen2025janus}  & 0.00 & \underline{10.52} & 2.03 & 16.12 & 4.15 & 23.53 & 7.21 & 10.13 & 9.68 & 21.32 & 2.43 & 3.40 & 9.21 \\

    Llama3.2-Vision \cite{lee2025efficient}  & 7.21 & 1.52 & 1.71 & 17.93 &  \textbf{22.80} & 29.08 & 0.00 & 12.21 & 11.98 & 15.33 & 7.13 & 11.82 & 11.56 \\

    Gemma3 \cite{gemmateam2025} & \textbf{14.85} & 10.50 & \textbf{27.63} & 20.11 & \underline{18.39} & \underline{43.32} & \underline{15.64} & \textbf{20.05} & \textbf{16.53} & \underline{38.82} & \underline{10.45} & \underline{19.43} & \underline{21.31} \\

    Qwen2.5-VL \cite{Qwen2.5-VL}  & 1.53 & 0.00 & 3.66 & \textbf{38.67} & 0.00 & 31.36 & 0.00 & 5.22 & 1.39 & 29.77 & 1.22 & 7.90 & 10.06 \\
    \bottomrule
    Focal-RegionFace & \underline{14.29} & \textbf{29.80} & \underline{17.24} & \underline{30.10} & 7.99 & \textbf{44.08} & \textbf{27.15} & \underline{14.70} & \underline{10.37} & \textbf{40.17} & \textbf{13.81} & \textbf{27.74} & \textbf{23.12} \\
    \bottomrule
\end{tabular}
\end{adjustbox}
\end{table*}

\subsection{Ablation Study: The Effect of the Multi-stage From I to III Detailed Breakdowns of Each Metric.} In Section 4.4, we conducted ablation studies from Stage I to Stage III to evaluate the model's performance at each stage. In this section, we provide additional details to complement the figures and tables presented in the main text. As shown in Table~\ref{tab:Ablation Study}, we report the detailed scores across five evaluation metrics and the win rates for all three stages. The results indicate that in Stage I, our model underperforms compared to other models across all metrics. However, after progressing to Stage II, there is a significant improvement in all scores, demonstrating the effectiveness and success of our region-aware face visual-language alignment strategy. In Stage III, the model’s performance further improves over Stage II, though the gain is not as dramatic. This is expected, as Stage III mainly focuses on enforcing box-level attention—masking forces the model to generate descriptions strictly based on the target region—which further validates the robustness and relevance of our design.

\begin{table*}[h!]
    \centering
    \caption{Detailed ablation studies based on the proposed MLLM-based evaluation.}
    \renewcommand{\arraystretch}{1.2} 
    \setlength{\tabcolsep}{8pt} 
    \resizebox{0.95\textwidth}{!}{
    \begin{tabular}{l|c|ccccc}
        \toprule
        & & \multicolumn{5}{c}{\textbf{GPT-4o} \cite{ray_2023_chatgpt}} \\ 
        \cmidrule(lr){0-2} \cmidrule(lr){3-7}
        & \textbf{Metrics} & Focal-RegionFace & Qwen2.5-VL \cite{Qwen2.5-VL} & Gemma3 \cite{gemmateam2025} & Deepseek-Janus-Pro \cite{chen2025janus} & Llama3.2-Vision \cite{lee2025efficient}\\ 
        \midrule
        \textbf{Stage I} & 
        \begin{tabular}[c]{@{}c@{}} \textbf{Cls} \\ \textbf{Det} \\ \textbf{Flu} \\ \textbf{Box} \\ \textbf{Sem} \\ \textbf{Win/\%} \end{tabular} & 
        \begin{tabular}[c]{@{}c@{}} 
            56.85 \\ 26.06 \\ 36.82 \\ 59.99 \\ 48.82 \\ 0.14 
        \end{tabular} &  
        \begin{tabular}[c]{@{}c@{}} 
            \underline{77.11} \\ \underline{77.68} \\ \textbf{84.98} \\ \textbf{82.43} \\ \underline{79.54} \\ \underline{41.28} 
        \end{tabular} & 
        \begin{tabular}[c]{@{}c@{}} 
            \textbf{77.96} \\ \textbf{79.94} \\ \underline{81.22} \\ \underline{80.97} \\ \textbf{81.05} \\ \textbf{49.96} 
        \end{tabular} & 
        \begin{tabular}[c]{@{}c@{}} 
            67.37 \\ 49.94 \\ 77.21 \\ 75.53 \\ 67.06 \\ 2.5 
        \end{tabular} & 
        \begin{tabular}[c]{@{}c@{}} 
            70.15 \\ 58.36 \\ 64.70 \\ 72.44 \\ 67.07 \\ 6.12 
        \end{tabular} \\ 
        \midrule
        \textbf{Stage II} & 
        \begin{tabular}[c]{@{}c@{}} \textbf{Cls} \\ \textbf{Det} \\ \textbf{Flu} \\ \textbf{Box} \\ \textbf{Sem} \\ \textbf{Win/\%} \end{tabular} & 
        \begin{tabular}[c]{@{}c@{}} \textbf{74.65} \\ \textbf{83.06} \\ \textbf{84.34} \\ \textbf{79.82} \\ \textbf{79.50} \\ \textbf{49.97} \end{tabular} &  
        \begin{tabular}[c]{@{}c@{}} \underline{67.28} \\ \underline{64.70} \\ \underline{78.17} \\ \underline{74.66} \\ \underline{69.64} \\ 18.09 \end{tabular} & 
        \begin{tabular}[c]{@{}c@{}} 67.2 \\ 60.16 \\ 72.18 \\ 72.83 \\ 68.04 \\ \underline{27.96} \end{tabular} & 
        \begin{tabular}[c]{@{}c@{}} 55.04 \\ 40.07 \\ 73.41 \\ 67.90 \\ 55.82 \\ 0.76 \end{tabular} & 
        \begin{tabular}[c]{@{}c@{}} 63.36 \\ 51.12 \\ 60.81 \\ 67.57 \\ 61.30 \\ 3.22 \end{tabular} \\
        \midrule
        \textbf{Stage III} & 
        \begin{tabular}[c]{@{}c@{}} \textbf{Cls} \\ \textbf{Det} \\ \textbf{Flu} \\ \textbf{Box} \\ \textbf{Sem} \\ \textbf{Win/\%} \end{tabular} & 
        \begin{tabular}[c]{@{}c@{}} \textbf{74.38} \\ \textbf{83.86} \\ \textbf{84.72} \\ \textbf{81.33} \\ \textbf{79.51} \\ \textbf{54.71} \end{tabular} &  
        \begin{tabular}[c]{@{}c@{}} 67.28 \\ \underline{64.62} \\ \underline{78.20} \\ \underline{74.73} \\ \underline{69.68} \\ 14.57 \end{tabular} & 
        \begin{tabular}[c]{@{}c@{}} \underline{67.35} \\ 60.10 \\ 72.15 \\ 71.70 \\ 68.16 \\ \underline{19.47} \end{tabular} & 
        \begin{tabular}[c]{@{}c@{}} 55.20\\ 39.91 \\ 73.41 \\ 68.19 \\ 55.98 \\ 7.55 \end{tabular} & 
        \begin{tabular}[c]{@{}c@{}} 63.61 \\ 51.16 \\ 60.78 \\ 67.85 \\ 61.53 \\ 3.70 \end{tabular} \\
        \bottomrule
    \end{tabular}
    }
    \label{tab:Ablation Study}
\end{table*}

\begin{table*}[h!]
    \centering
    \caption{Detailed ablation studies based on the proposed MLLM-based evaluation.}
    \renewcommand{\arraystretch}{1.2} 
    \setlength{\tabcolsep}{8pt} 
    \resizebox{0.95\textwidth}{!}{
    \begin{tabular}{l|c|ccccc}
        \toprule
        & & \multicolumn{5}{c}{\textbf{GPT-4o} \cite{ray_2023_chatgpt}} \\ 
        \cmidrule(lr){0-2} \cmidrule(lr){3-7}
        & \textbf{Metrics} & Focal-RegionFace & Qwen2.5-VL \cite{Qwen2.5-VL} & Gemma3 \cite{gemmateam2025} & Deepseek-Janus-Pro \cite{chen2025janus} & Llama3.2-Vision \cite{lee2025efficient}\\ 
        \midrule
        \textbf{Single Stage} & 
        \begin{tabular}[c]{@{}c@{}} \textbf{Cls} \\ \textbf{Det} \\ \textbf{Flu} \\ \textbf{Box} \\ \textbf{Sem} \\ \textbf{Win/\%} \end{tabular} & 
        \begin{tabular}[c]{@{}c@{}} 
            65.47 \\ 58.24 \\ 38.57 \\ 73.22 \\ 68.83 \\ 23.80 
        \end{tabular} &  
        \begin{tabular}[c]{@{}c@{}} 
            \textbf{75.47} \\ \underline{74.96} \\ \textbf{79.89} \\ \textbf{81.52} \\ \underline{74.42} \\ \underline{32.72} 
        \end{tabular} & 
        \begin{tabular}[c]{@{}c@{}} 
            \underline{73.12} \\ \textbf{71.55} \\ \underline{77.51} \\ \underline{78.94} \\ \textbf{80.72} \\ \textbf{37.24} 
        \end{tabular} & 
        \begin{tabular}[c]{@{}c@{}} 
            59.38 \\ 44.31 \\ 73.96 \\ 68.42 \\ 59.69 \\ 2.50 
        \end{tabular} & 
        \begin{tabular}[c]{@{}c@{}} 
            67.43 \\ 55.23 \\ 62.77 \\ 70.82 \\ 58.86 \\ 3.74 
        \end{tabular} \\
        \bottomrule
    \end{tabular}
    }
    \label{tab:Ablation Study2}
\end{table*}

\begin{table*}[h!]
    \centering
    \caption{Detailed ablation studies based on the proposed MLLM-based evaluation.}
    \renewcommand{\arraystretch}{1.2} 
    \setlength{\tabcolsep}{8pt} 
    \resizebox{0.95\textwidth}{!}{
    \begin{tabular}{l|c|ccccc}
        \toprule
        & & \multicolumn{5}{c}{\textbf{GPT-4o} \cite{ray_2023_chatgpt}} \\ 
        \cmidrule(lr){0-2} \cmidrule(lr){3-7}
        & \textbf{Metrics} & Focal-RegionFace & Qwen2.5-VL \cite{Qwen2.5-VL} & Gemma3 \cite{gemmateam2025} & Deepseek-Janus-Pro \cite{chen2025janus} & Llama3.2-Vision \cite{lee2025efficient}\\ 
        \midrule
        \textbf{Stage III} & 
        \begin{tabular}[c]{@{}c@{}} \textbf{Cls} \\ \textbf{Det} \\ \textbf{Flu} \\ \textbf{Box} \\ \textbf{Sem} \\ \textbf{Win/\%} \end{tabular} & 
        \begin{tabular}[c]{@{}c@{}} 
            20.73 \\ 53.56 \\ 73.42 \\ 53.20 \\ 67.53 \\ 11.67 
        \end{tabular} &  
        \begin{tabular}[c]{@{}c@{}} 
            \underline{78.41} \\ \textbf{79.33} \\ \underline{81.48} \\ \underline{78.90} \\ \underline{75.89} \\ \underline{38.64} 
        \end{tabular} & 
        \begin{tabular}[c]{@{}c@{}} 
            \textbf{79.13} \\ \underline{78.63} \\ \textbf{83.44} \\ \textbf{81.69} \\ \textbf{80.65} \\ \textbf{41.60} 
        \end{tabular} & 
        \begin{tabular}[c]{@{}c@{}} 
            65.45 \\ 47.41 \\ 74.23 \\ 71.78 \\ 69.32 \\ 3.20 
        \end{tabular} & 
        \begin{tabular}[c]{@{}c@{}} 
            70.84 \\ 60.19 \\ 63.42 \\ 70.77 \\ 68.33 \\ 4.89 
        \end{tabular} \\ 
        \midrule
        \textbf{Stage I+III} & 
        \begin{tabular}[c]{@{}c@{}} \textbf{Cls} \\ \textbf{Det} \\ \textbf{Flu} \\ \textbf{Box} \\ \textbf{Sem} \\ \textbf{Win/\%} \end{tabular} & 
        \begin{tabular}[c]{@{}c@{}} 66.48 \\ \textbf{68.47} \\ 60.87 \\ 71.46 \\ \textbf{70.23} \\ 25.42 \end{tabular} &  
        \begin{tabular}[c]{@{}c@{}} \textbf{69.41} \\ 67.84 \\ \textbf{77.23} \\ \textbf{76.91} \\ \underline{70.01} \\ \underline{28.31} \end{tabular} & 
        \begin{tabular}[c]{@{}c@{}} \underline{68.23} \\ \underline{67.85} \\ \underline{72.87} \\ \underline{76.42} \\ 69.10 \\ \textbf{29.94} \end{tabular} & 
        \begin{tabular}[c]{@{}c@{}} 54.26 \\ 41.98 \\ 69.38 \\ 68.50 \\ 60.41 \\ 6.99 \end{tabular} & 
        \begin{tabular}[c]{@{}c@{}} 64.41 \\ 55.26 \\ 66.85 \\ 71.43 \\ 65.43 \\ 9.34 \end{tabular} \\
        \midrule
        \textbf{Stage II+III} & 
        \begin{tabular}[c]{@{}c@{}} \textbf{Cls} \\ \textbf{Det} \\ \textbf{Flu} \\ \textbf{Box} \\ \textbf{Sem} \\ \textbf{Win/\%} \end{tabular} & 
        \begin{tabular}[c]{@{}c@{}} \textbf{71.56} \\ \textbf{80.31} \\ \textbf{82.44} \\ \textbf{90.07} \\ \textbf{81.08} \\ \textbf{49.51} \end{tabular} &  
        \begin{tabular}[c]{@{}c@{}} 63.26 \\ \underline{61.33} \\ \underline{78.43} \\ 72.76 \\ \underline{72.54} \\ 19.23 \end{tabular} & 
        \begin{tabular}[c]{@{}c@{}} \underline{65.44} \\ 60.58 \\ 67.65 \\ \underline{74.70} \\ 68.42 \\ \underline{20.21} \end{tabular} & 
        \begin{tabular}[c]{@{}c@{}} 63.78 \\ 59.26 \\ 66.42 \\ 52.17 \\ 57.32 \\ 8.23 \end{tabular} & 
        \begin{tabular}[c]{@{}c@{}} 61.33 \\ 60.71 \\ 61.28 \\ 70.14 \\ 47.16 \\ 2.82 \end{tabular} \\
        \bottomrule
    \end{tabular}
    }
    \label{tab:Ablation Study3}
\end{table*}

\subsection{Ablation Study: Comparing Multi-Stage and Single-Stage Fine-Tuning} We evaluate the model under single-stage fine-tuning using the same experimental setup as in previous sections, reporting five evaluation metrics and win rates in Table~\ref{tab:Ablation Study2}. As shown, single-stage fine-tuning consistently underperforms the multi-stage approach across all metrics. This gap indicates that jointly training on heterogeneous data and objectives in a single stage impedes effective knowledge acquisition. In contrast, the multi-stage strategy enforces a structured learning progression, enabling the model to acquire foundational representations before addressing more complex tasks, which leads to more robust and generalizable performance.

\subsection{Ablation Study: Selective Fine-Tuning Combinations.} To analyze the contribution of each fine-tuning stage, we evaluate three partial configurations: Base + Stage III, Base + Stage I + Stage III, and Base + Stage II + Stage III (Table~\ref{tab:Ablation Study3}). Training with only Stage III leads to severe performance degradation, as the model lacks both global perception and region-aware alignment, indicating that region-masked fine-tuning depends on prior region semantic learning. Incorporating Stage I improves results but remains unstable without Stage II, showing that global perception alone cannot replace region-aware supervision. In contrast, Base + Stage II + Stage III achieves strong performance, confirming Stage II’s central role in establishing localized semantic understanding, though it still slightly underperforms the full pipeline. Overall, these results demonstrate that progressive fine-tuning is essential: Stage II is indispensable for enabling Stage III, while Stage I consistently enhances robustness and generalization.

\section{Visualisation More Examples}
\begin{figure}[H]
  \centering
  \includegraphics[width=1\linewidth]{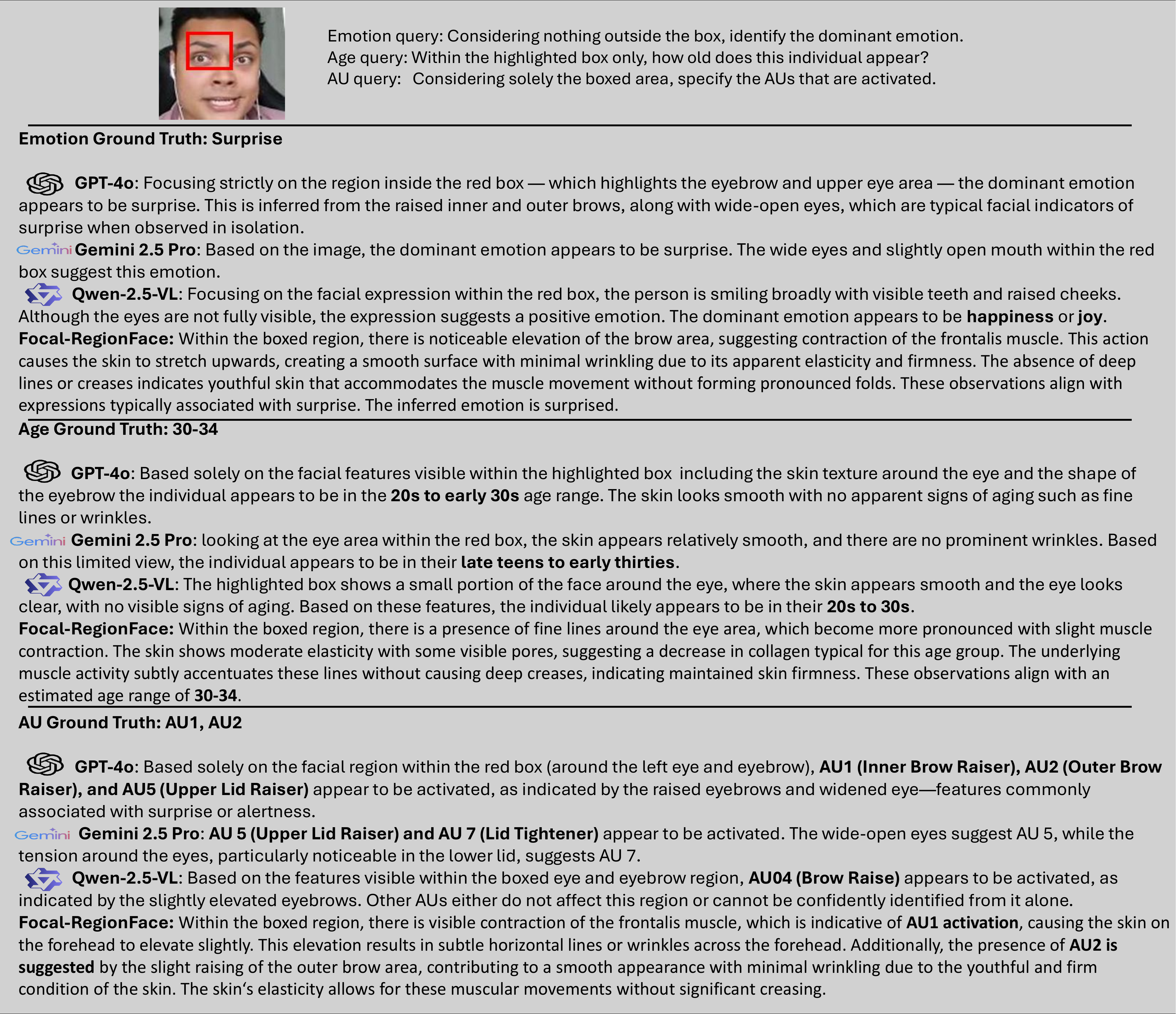}
  \caption{Visualisation Example 1.}
  \label{fig:Visualisation Example 1}
\end{figure}

\bibliographystyle{elsarticle-num} 
\bibliography{custom}










\end{document}